\pdfoutput=1
\documentclass[11pt]{article}

\usepackage[]{acl}

\usepackage{times}
\usepackage{latexsym}

\usepackage[T1]{fontenc}
\usepackage[utf8]{inputenc}

\usepackage{microtype}

\usepackage{amsmath}
\usepackage{pifont}
\DeclareMathOperator*{\argmin}{argmin} % thin space, limits underneath in displays
\newcommand{\cmark}{\ding{51}}%
\usepackage{booktabs}
\usepackage{blindtext}
\usepackage{multirow}
\usepackage[para,online,flushleft]{threeparttable}
\usepackage{algorithm}
\usepackage{algpseudocode}

\usepackage{graphicx}
\usepackage{subfigure} 
\usepackage{enumitem}
\usepackage{tcolorbox}
\newcommand{\ie}{\emph{i.e.,}\xspace}

\title{Few-shot Event Detection: An Empirical Study and a Unified View}

\author{
 Yubo~Ma$^{1}$, Zehao~Wang$^{2}$, Yixin Cao$^{3\dag}$, Aixin Sun$^{1\dag}$ \\ 
 \\
 $^1$ S-Lab, Nanyang Technological University \\
 $^2$ KU Leuven
 $^3$ Singapore Management University \\
\texttt{yubo001@e.ntu.edu.sg}\\
}

\begin{document}
\maketitle
\renewcommand{\thefootnote}{\fnsymbol{footnote}}
\footnotetext[2]{Corresponding Author.}
\renewcommand{\thefootnote}{\arabic{footnote}}

\begin{abstract}
Few-shot event detection (ED) has been widely studied, while this brings noticeable discrepancies, e.g., various motivations, tasks, and experimental settings, that hinder the understanding of models for future progress.
This paper presents a thorough empirical study, a unified view of ED models, and a better \textit{unified baseline}. For fair evaluation, we compare 12 representative methods on three datasets, which are roughly grouped into prompt-based and prototype-based models for detailed analysis. Experiments consistently demonstrate that prompt-based methods, including ChatGPT, still significantly trail prototype-based methods in terms of overall performance. To investigate their superior performance, we break down their design elements along several dimensions and build a unified framework on prototype-based methods. Under such unified view, each prototype-method can be viewed a combination of different modules from these design elements. We further combine all advantageous modules and propose a simple yet effective \textit{baseline}, which outperforms existing methods by a large margin (e.g., $2.7\%$ $F1$ gains under \textit{low-resource} setting).~\footnote{Our code will be publicly available at \texttt{https://github.com/mayubo2333/fewshot\_ED}.}
\end{abstract}

%==============================
\section{Introduction}
%==============================

Event Detection (ED) is the task of identifying event triggers and types in texts. For example, given \textit{``Cash-strapped Vivendi wants to \underline{sell} Universal Studios''}, it is to classify the word \textit{``sell''} into a \textit{TransferOwnership} event. ED is a fundamental step in various tasks such as successive event-centric information extraction~\cite{huang-etal-2022-multilingual-generative, ma-etal-2022-prompt, chen-etal-2022-ergo}, knowledge systems~\cite{li-etal-2020-gaia, wen-etal-2021-resin}, story generation~\cite{li-etal-2022-event}, etc. However, the annotation of event instances is costly and labor-consuming, which motivates the research on improving ED with limited labeled samples, i.e., the few-shot ED task. 

\begin{figure}
    \centering
    \includegraphics[width=\linewidth]{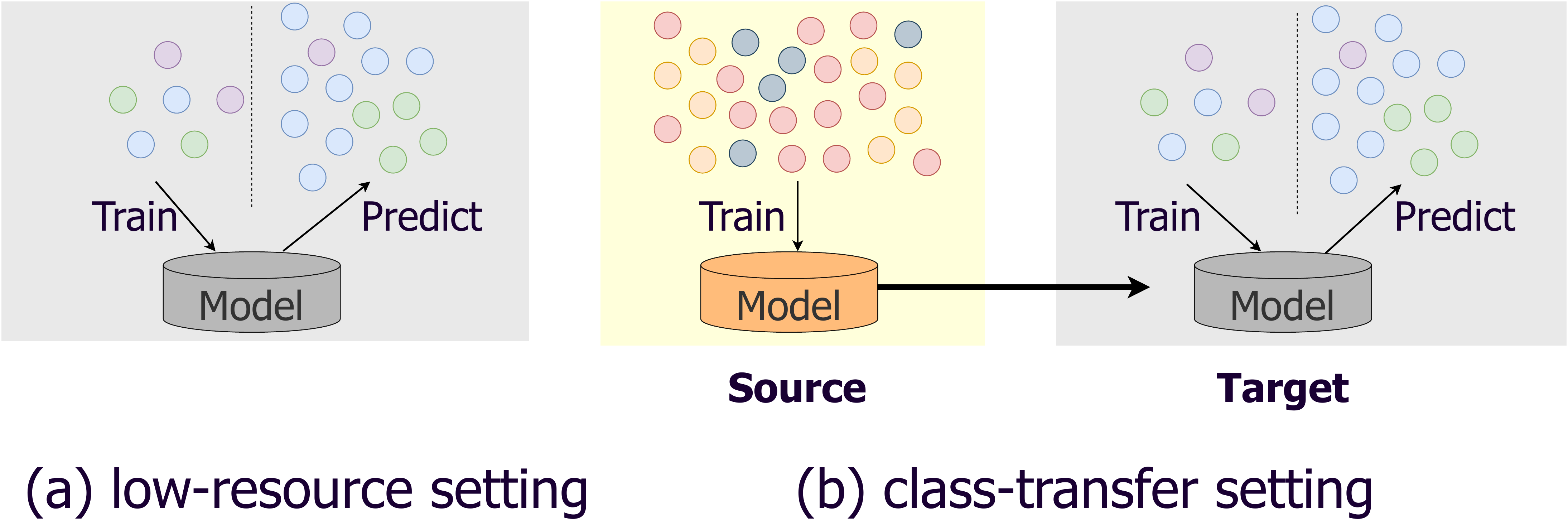}
    \caption{
        Task settings to access \textit{Generalization} (a) and \textit{Transferability} (b). Colors denote event types.
    }
    \label{fig:two definitions}
\end{figure}

Extensive studies have been carried out on few-shot ED. Nevertheless, there are noticeable discrepancies among existing methods from three aspects. 
(1) \textit{Motivation} (Figure~\ref{fig:two definitions}): Some methods focus on model's \textit{generalization} ability that learns to classify with only a few samples~\cite{Li_2022_PILED}. Some other methods improve the \textit{transferability}, by introducing additional data, that adapts a well-trained model on the preexisting schema to a new schema using a few samples~\cite{lu-etal-2021-text2event}. There are also methods considering both abilities~\cite{liu-etal-2020-event, hsu-etal-2022-degree}. (2) \textit{Task setting}: Even focusing on the same ability,  methods might adopt different task settings for training and evaluation. For example, there are at least three settings for transferability: \textit{episode learning} (EL, \citealt{Deng_2020, cong-etal-2021-shot}), \textit{class-transfer} (CT, \citealt{hsu-etal-2022-degree}) and \textit{task-transfer} (TT, \citealt{lyu-etal-2021-zero, lu-etal-2022-unified}). (3) \textit{Experimental Setting}: Even focusing on the same task setting, their experiments may vary in different sample sources (e.g., a subset of datasets, annotation guidelines, or external corpus) and sample numbers (shot-number or sample-ratio). Table~\ref{tab:method comparison} provides a detailed comparison of representative methods.

\begin{table*}
\centering
\small
\setlength\tabcolsep{3pt}
\caption{
Noticeable discrepancies among existing few-shot ED methods. Explanations of task settings can be found in Section~\ref{subsec:task setting}, which also refer to different motivations: LR for generalization, EL, CT, and TT for transfer abilities.  \textbf{Dataset} indicates the datasets on which the training and/or evaluation is conducted. \textbf{Sample Number} refers to the number of labeled samples used. \textbf{Sample Source} refers to where training samples come from. Guidelines: example sentences from annotation guidelines. Datasets: subsets of full datasets. Corpus: (unlabeled) external corpus.
} 
\label{tab:method comparison}       
    \begin{tabular}{@{}ll|cccc|ccc} 
    \toprule      &Method & \multicolumn{4}{c|}{\textbf{Task setting}}  &  \multicolumn{3}{c}{\textbf{Experimental setting}} \\
    & & LR& EL&CT&TT & Dataset & Sample Number & Sample Source \\
    \midrule
     \parbox[t]{2mm}{\multirow{10}{*}{\rotatebox[origin=c]{90}{Prototype-based}}}
    & \shortstack{Seed-based~\scriptsize{\cite{bronstein-etal-2015-seed}}} &  &  & \cmark & & ACE & 30 & Guidelines \\
    & \shortstack{MSEP~\scriptsize{\cite{peng-etal-2016-event}}}  & \cmark &  & \cmark &  &   ACE  & 0 & Guidelines  \\
    &\shortstack{ZSL~\scriptsize{\cite{huang-etal-2018-zero}}}  &  &  & \cmark &  & ACE & 0 & Datasets  \\
    & \shortstack{DMBPN~\scriptsize{\cite{Deng_2020}}} &  & \cmark & &  & FewEvent &  \{5,10,15\}-shot   & Datasets  \\
    & \shortstack{OntoED~\scriptsize{\cite{deng-etal-2021-ontoed}}} & \cmark &  & \cmark &  & MAVEN / FewEvent &
    \{0,1,5,10,15,20\}\%  & Datasets  \\
    & \shortstack{Zhang's~\scriptsize{\cite{zhang-etal-2021-zero}}} & \cmark &  &    &  & ACE & 0 & Corpus \\ 
    & \shortstack{PA-CRF~\scriptsize{\cite{cong-etal-2021-shot}}} & & \cmark &  &   & FewEvent & \{5,10\}-shot & Datasets \\
    & \shortstack{ProAcT~\scriptsize{\cite{lai-etal-2021-learning}}} &  & \cmark &   &  & ACE / FewEvent / RAMS  & \{5,10\}-shot & Datasets \\
    & \shortstack{CausalED~\scriptsize{\cite{chen-etal-2021-honey}}} &  & \cmark &  &  & ACE / MAVEN / ERE & 5-shot & Datasets \\
    & \shortstack{Yu's~\scriptsize{\cite{yu-etal-2022-building}}} & \cmark &  &  &   & ACE  & 176 & Guidelines + Corpus \\
    & \shortstack{ZED~\scriptsize{\cite{zhang-etal-2022-efficient-zero}}} & \cmark  &  &  &   & MAVEN &  0 & Corpus \\
    & \shortstack{HCL-TAT~\scriptsize{\cite{zhang-etal-2022-hcl}}} &  & \cmark &  &   & FewEvent  &  \{5,10\}-shot & Datasets  \\
    & \shortstack{KE-PN~\scriptsize{\cite{zhao-etal-2022-knowledge}}} &  & \cmark &  &   & ACE / MAVEN / FewEvent  &  \{1,5\}-shot & Datasets \\
    \midrule
    \parbox[t]{2mm}{\multirow{7}{*}{\rotatebox[origin=c]{90}{Prompt-based}}}
    & \shortstack{EERC~\scriptsize{\cite{liu-etal-2020-event}}} & \cmark &  & \cmark & \cmark & ACE &  \{0,1,5,10,20\}\% & Datasets  \\
    & \shortstack{FSQA~\scriptsize{\cite{2020_Feng}}}  & \cmark &  & & \cmark & ACE & \{0,1,3,5,7,9\}-shot & Datasets \\ 
    & \shortstack{EDTE~\scriptsize{\cite{lyu-etal-2021-zero}}}  &  &  &  & \cmark & ACE / ERE & 0 & -   \\ 
    & \shortstack{Text2Event~\scriptsize{\cite{lu-etal-2021-text2event}}}  &  &  & \cmark &   & ACE / ERE& \{1,5,25\}\% & Datasets \\
    & \shortstack{UIE~\scriptsize{\cite{lu-etal-2022-unified}}}  & \cmark &  & \cmark &   & ACE / CASIE  &  \{1,5,10\}-shot/\% & Datasets \\
    & \shortstack{DEGREE~\scriptsize{\cite{hsu-etal-2022-degree}}}  & \cmark & & \cmark &  & ACE / ERE  &  \{0,1,5,10\}-shot & Datasets \\
    & \shortstack{PILED~\scriptsize{\cite{Li_2022_PILED}}}   & \cmark & \cmark &  &  & ACE / MAVEN / FewEvent  &  \{5,10\}-shot & Datasets\\
    \bottomrule 
    \end{tabular}
\end{table*}

In this paper, we argue the importance of a unified setting for a better understanding of few-shot ED. First, based on exhaustive background investigation on ED and similar tasks (e.g., NER), we conduct \textbf{an empirical study of twelve SOTA methods under two practical settings}: \textit{low-resource} setting for \textit{generalization} ability and  \textit{class-transfer} setting for \textit{transferability}. 
We roughly classify the existing methods into two groups: prototype-based models to learn event-type representations and proximity measurement for prediction and prompt-based models that convert ED into a familiar task of Pre-trained Language Models (PLMs).

The second contribution is \textbf{a unified view of prototype-based methods} to investigate their superior performance. Instead of picking up the best-performing method as in conventional empirical studies, we take one step further. We break down the design elements along several dimensions, e.g., the source of prototypes, the aggregation form of prototypes, etc. From this perspective, five prototype-based methods on which we conduct experiment are instances of distinct modules from these elements. And third, through analyzing each effective design element, we propose \textbf{a simple yet effective \textit{unified baseline}} that combines all advantageous elements of existing methods. Experiments validate an average $2.7\%$ $F1$ gains under \textit{low-resource} setting and the best performance under \textit{class-transfer} setting. Our analysis also provides many valuable insights for future research.
%============================
\section{Preliminary}
%============================

Event detection (ED) is usually formulated as either a span classification task or a sequence labeling task, depending on whether candidate event spans are provided as inputs.
We brief the sequence labeling paradigm here because the two paradigms can be easily converted to each other.

Given a dataset $\mathcal{D}$ annotated with schema $E$ (the set of event types) and a sentence $X = [x_1, ..., x_N]^T \in \mathcal{D}$, where $x_i$ is the $i$-th word and $N$ the length of this sentence, ED aims to assign a label $y_i \in \left( E \cup \{\texttt{N.A.}\}\right)$ for each $x_i$ in $X$. Here \texttt{N.A.} refers to either none events or events beyond pre-defined types $E$. We say that word $x_i$ triggering an event $y_i$ if $y_i \in E$.

%======================
\subsection{Few-shot ED task settings}
\label{subsec:task setting}
%======================
We categorize few-shot ED settings to four cases: \textit{low-resource} (LR), \textit{class-transfer} (CT), \textit{episode learning} (EL) and \textit{task-transfer} (TT). Low-resource setting assesses the \textit{generalization} ability of few-shot ED methods, while the other three settings are for \textit{transferability}. We adopt LR and CT in our empirical study towards practical scenarios. More details can be found in Appendix~\ref{subsec:four settings}.

\noindent{\textbf{Low-resource setting}} assumes access to a dataset $\mathcal{D} = (\mathcal{D}_{train}, \mathcal{D}_{dev}, \mathcal{D}_{test})$ annotated with a label set $E$, where  $|\mathcal{D}_{dev}| \leq |\mathcal{D}_{train}| \ll |\mathcal{D}_{test}|$. It assesses the generalization ability of models by (1) utilizing only few samples during training, and (2) evaluating on the real and rich test dataset. 

\noindent{\textbf{Class-transfer setting}} assumes access to a source dataset $\mathcal{D}^{(S)}$ with a preexisting schema $E^{(S)}$ and a target dataset $\mathcal{D}^{(T)}$ with a new schema $E^{(T)}$. Note that $D^{(S)}$ and $D^{(T)}$, $E^{(S)}$ and $E^{(T)}$ contain disjoint sentences and event types, respectively. $\mathcal{D}^{(S)}$ contains abundant samples, while $\mathcal{D}^{(T)}$ is the low-resource setting dataset described above. Models under this setting are expected to be pre-trained on $\mathcal{D}^{(S)}$ then further trained and evaluated on $\mathcal{D}^{(T)}$.

%============================
\subsection{Category of existing methods}
\label{subsec:approach}
%============================
We roughly group existing few-shot ED methods into two classes: prompt-based methods and prototype-based methods. More details are introduced in Appendix~\ref{subsec:method appendix}.

\noindent{\textbf{Prompt-based methods}} leverage the rich language knowledge in PLMs by converting downstream tasks to the task with which PLMs are more familiar. 
Such format conversion narrows the gap between pre-training and downstream tasks and benefits knowledge induction in PLMs with limited annotations. Specifically, few-shot ED can be converted to machine reading comprehension (MRC,~\citealt{du-cardie-2020-event, liu-etal-2020-event, 2020_Feng}), natural language inference (NLI,~\citealt{lyu-etal-2021-zero}), conditional generation (CG,~\citealt{2021_Paolini, lu-etal-2021-text2event, lu-etal-2022-unified, hsu-etal-2022-degree}), and the cloze task~\cite{Li_2022_PILED}. We give examples of these prompts in Table~\ref{tab: diff prompt}.

\noindent{\textbf{Prototype-based methods}}
predict an event type for each word/span mention by measuring its representation proximity to \textit{prototypes}.
Here we define prototypes in a \textit{generalized} format --- it is an embedding that represents some event type.
For example, Prototypical Network (ProtoNet,~\citealt{Snell_2017}) and its variants~\cite{Lai_2020, lai-etal-2020-extensively, Deng_2020, deng-etal-2021-ontoed, cong-etal-2021-shot, lai-etal-2021-learning} construct prototypes via a subset of sample mentions. In addition to event mentions, a line of work leverage related knowledge to learn or enhance prototypes' representation, including AMR graphs~\cite{huang-etal-2018-zero}, event-event relations~\cite{deng-etal-2021-ontoed}, definitions~\cite{shen-etal-2021-adaptive} and FrameNet~\cite{zhao-etal-2022-knowledge}. \citet{zhang-etal-2022-hcl} recently introduce contrastive learning~\cite{2006_CL} in few-shot ED task. Such method also determines the event by measuring the distances with other samples and aggregates these distances to evaluate an overall distance to each event type. Therefore we view it as a \textit{generalized} format of prototype-based methods as well.

For comprehensiveness, we also include competitive methods from similar tasks, \ie Named Entity Recognition and Slot Tagging, which are highly adaptable to ED. Such expansion enriches the categorization and enables us to build a unified view in Section~\ref{sec:unified view}. For instance, some methods~\cite{hou-etal-2020-shot, ma-etal-2022-label} leverage label semantics to enhance or directly construct the prototypes. Others~\cite{das-etal-2022-container} leverage contrastive learning for better prototype representations.
\section{A Prototype-based Unified View}
\label{sec:unified view}

\begin{figure}
    \includegraphics[width=\linewidth]{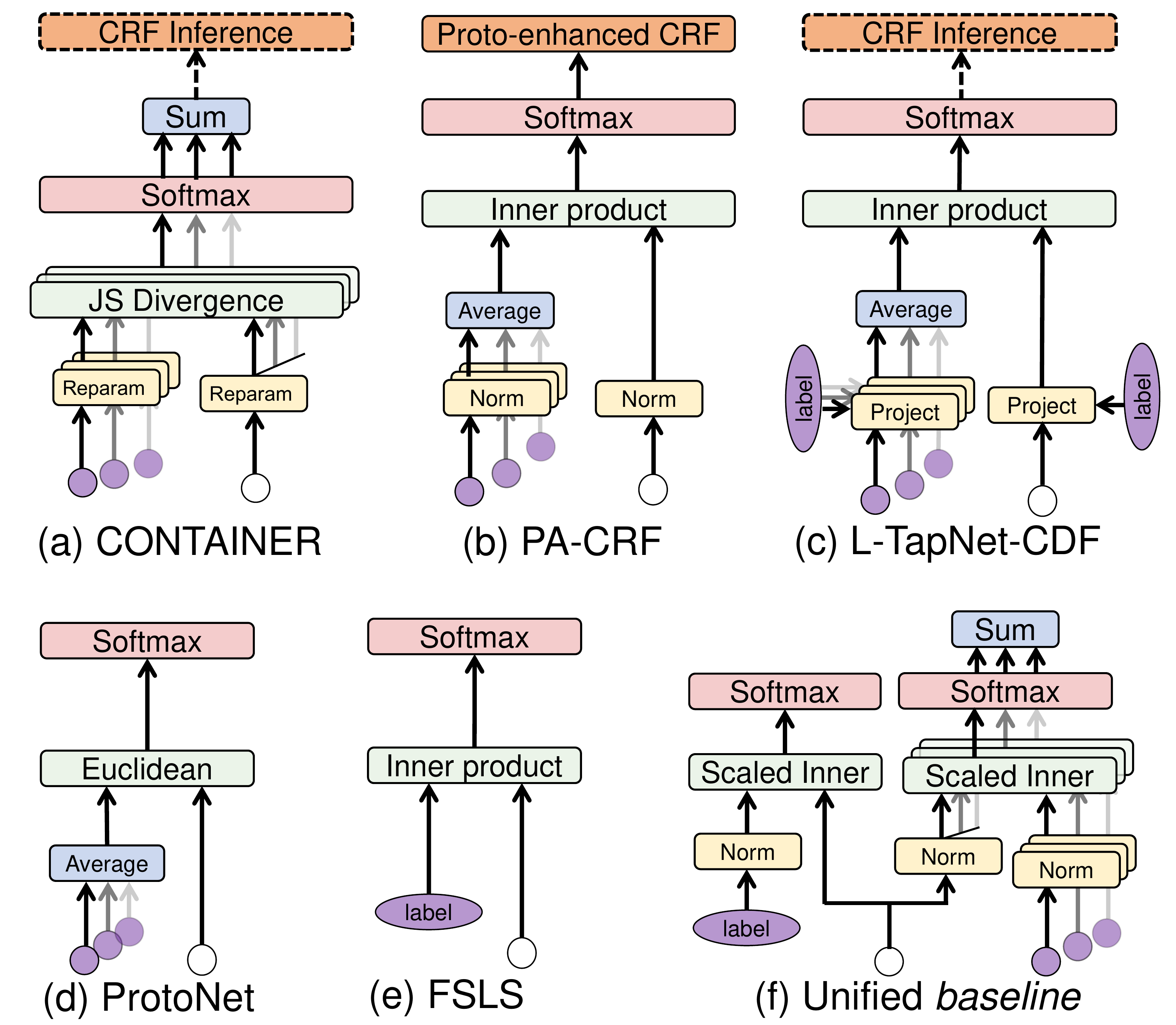}
    \caption{The architectures of five existing prototype-based methods and the unified baseline. Given event mention $x$ and event type $y$, each sub-figure depicts how to compute the $\text{logits}(y|x)$. White circles: representation of predicted event $h_x$. Purple circles: representation of prototypes $h_{c_y}$ ($c_y \in \mathcal{C}_y$). Yellow modules: transfer functions. Green modules: distance functions. Blue modules: aggregation form. Orange modules: CRF modules. Dashed lines in (a) and (c) represent that their CRFs are only used during inference.}
    \label{fig:model arch}
\end{figure}

\begin{table*}[t]
\centering
\small
\caption{Decomposing five prototype-based methods and \textit{unified baseline} along design elements. "Both" in column 1 means both event mentions and label names for $y$ are prototype sources. JSD: Jensen–Shannon divergence. $\mathcal{M}$: Projection matrix in TapNet. $\mathcal{N}(\mu(h), \Sigma(h))$: Gaussian distribution with mean $\mu(h)$ and covariance matrix $\Sigma(h)$.
} 
\label{tab:universal framework}  
\begin{tabular}{@{}l|ccccc} 
    \toprule
 Method & Prototype $\mathcal{C}_y$ & Aggregation &Distance $d(u, v)$ & Transfer $f(h)$ & CRF Module   \\
    \midrule
    ProtoNet~\scriptsize{\cite{Snell_2017}} & Event mentions & feature & $||u - v||_2$ & $h$ & $-$\\
    L-TapNet-CDT~\scriptsize{\cite{hou-etal-2020-shot}} & Both & feature & $-u^Tv/\tau$ & $\mathcal{M} \frac{h}{||h||}$& CRF-Inference \\
    PA-CRF~\scriptsize{\cite{cong-etal-2021-shot}} & Event mentions & feature & $-u^Tv$ & $\frac{h}{||h||}$ & CRF-PA \\
    CONTAINER~\scriptsize{\cite{das-etal-2022-container}} & Event mentions & score & $\text{JSD}(u||v)$ & $\mathcal{N}(\mu(h), \Sigma(h))$ & CRF-Inference \\
    FSLS~\scriptsize{\cite{ma-etal-2022-label}} & Label name & $-$ & $-u^Tv$ & $h$ & $-$ \\
    \midrule
    Unified Baseline (Ours) & Both & score + loss & $-u^Tv/\tau$ & $\frac{h}{||h||}$ & $-$ \\
    \bottomrule 
    \end{tabular}
\end{table*}

Due to the superior performance (Sections~\ref{sec:low-resource learning} and \ref{sec:class-transfer learning}), we zoom into prototype-based methods to provide a unified view towards a better understanding. We observe that they share lots of similar components. As shown in Table~\ref{tab:universal framework} and Figure~\ref{fig:model arch}, we decompose prototype-based methods into 5 design elements: prototype source, transfer function, distance function, aggregation form, and CRF module. This unified view enables us to compare choices in each design element directly. By aggregating the effective choices, we end with a \textit{Unified Baseline}.  

Formally, given an event mention $x$, prototype-based methods predict the likelihood $p(y|x)$ from $\text{logits}(y|x)$ for each $y \in (E \cup \{\texttt{N.A.}\})$
\begin{equation}
\nonumber
    p(y|x) = \text{Softmax}_{y \sim (E \cup \{\texttt{N.A.\}})} \text{logits}(y|x)
\end{equation}

The general framework is as follows. Denote the PLM's output representation of event mention $x$ and data $c_y$ in \underline{prototype source} $\mathcal{C}_y$ as $h_x$ and $h_{c_y}$ respectively, where $h \in R^m$ and $m$ is the dimension of PLM's hidden space. The first step is to convert $h_x$ and $h_{c_y}$ to appropriate representations via a \underline{transfer function} $f(\cdot)$. Then the methods maintain either a single or multiple prototypes $c_y$'s for each event type, determined by the adopted \underline{aggregation form}. Third, the distance between $f(h_x)$ and $f(h_{c_y})$ (single prototype) or $f(h_{c_y})$'s (multiple prototypes) is computed via a \underline{distance function} $d(\cdot, \cdot)$ to learn the proximity scores, \ie $\text{logits}(y|x)$. Finally, an optional \underline{CRF module} is used to adjust $\text{logits}(y|x)$ for $x$ in the same sentence to model their label dependencies. For inference, we adopt nearest neighbor classification by assigning the sample with nearest event type in $\cup_{y \in (E \cup \{\texttt{N.A.\}})} \mathcal{C}_y$ , \ie
\begin{equation}
\nonumber
    \hat{y}_x = \argmin_{y \in (E \cup \{\texttt{N.A.\}})} \min_{c_y \in \mathcal{C}_y} d(f(h_x), f(h_{c_y}))
\end{equation}

\noindent{Next, we detail the five design elements:}

\noindent{\textbf{Prototype source}} $\mathcal{C}_y$ (purple circles in Figure~\ref{fig:model arch}, same below) indicates a set about the source of data / information for constructing the prototypes. There are mainly two types of sources:

\noindent(1) \textit{event mentions} (purple circle without words): ProtoNet and its variants in Figure~\ref{fig:model arch}(b),(c),(d) additionally split a support set $\mathcal{S}_y$ from training data as prototype source, while contrastive learning methods in Figure~\ref{fig:model arch}(a) view every annotated mention as the source (except the query one).

\noindent(2) \textit{Label semantics} (purple ellipses with words): Sometimes, the label name $l_y$ is utilized as the source to enhance or directly construct the prototypes. For example, FSLS in Figure~\ref{fig:model arch}(e) views the text representation of type names as prototypes, while L-TapNet-CDT in Figure~\ref{fig:model arch}(c) utilizes both the above kinds of prototype sources.

\noindent{\textbf{Transfer function}} $f: R^m \rightarrow R^n$ (yellow modules) transfers PLM outputs into the distance space for prototype proximity measurement. Widely used transfer functions include normalization in Figure~\ref{fig:model arch}(b), down-projection in Figure~\ref{fig:model arch}(c), reparameterization in Figure~\ref{fig:model arch}(a), or an identity function.

\noindent{\textbf{Distance function}}  $d: R^n \times R^n \rightarrow R_+$ (green modules) measures the distance of two transferred representations within the same embedded space. Common distance functions are euclidean distance in Figure~\ref{fig:model arch}(d) and negative cosine similarity in Figure~\ref{fig:model arch}(b),(c),(e).

\noindent{\textbf{Aggregation form}} (blue modules) describes how to compute $\text{logits}(y|x)$ based on a single or multiple prototype sources. Aggregation may happen at three levels. 

\noindent{(1) \textit{feature-level}}: ProtoNet and its variants in Figure~\ref{fig:model arch}(b),(c),(d) aims to  construct a \textit{single} prototype $h_{\Bar{c}_y}$ for each event type $y$ by merging various features, which ease the calculation $\text{logits}(y|x) = -d(f(h_x), f(h_{\Bar{c}_y}))$.

\noindent{(2) \textit{score-level}}: CONTAINER in Figure~\ref{fig:model arch}(a) views each data as a prototype (they have \textit{multiple} prototypes for each type $y$) and computes the distance $d(f(h_x), f(h_{c_y}))$ for each $c_y \in \mathcal{C}_y$. These distances are then merged to obtain $\text{logits}(y|x)$. 

\noindent{(3) \textit{loss-level}}: Such form has multiple parallel branches $b$ for each mention $x$. Each branch has its own $\text{
logits}^{(b)}(y|x)$ and is optimized with different loss components during training. Thus it could be viewed as a multi-task learning format. See  \textit{unified baseline} in Figure~\ref{fig:model arch}(f).

\noindent{\textbf{CRF module}} (orange modules) adjusts predictions within the same sentence by explicitly considering the label dependencies between sequential inputs. The vanilla CRF~\cite{2001_CRF} and its variants in Figure~\ref{fig:model arch}(a),(b),(c) post additional constraints into few-shot learning.
\section{Experimental setup}

\subsection{Few-shot datasets and Evaluation}
\noindent{\textbf{Dataset source}.} 
We utilize ACE05~\cite{doddington-etal-2004-automatic}, MAVEN~\cite{wang-etal-2020-maven} and ERE~\cite{song-etal-2015-light} to construct few-shot ED datasets in this empirical study. Detailed statistics about these three datasets are in Appendix~\ref{subsec:full dataset}.

\noindent{\textbf{Low-resource setting}.} We adopt $K$-shot sampling strategy to construct few-shot datasets for the low-resource setting, i.e., sampling $K_{train}$ and $K_{dev}$ samples per event type to construct the train and dev sets, respectively.\footnote{Recent systematic research on few-shot NLP tasks~\cite{perez2021true} is of opposition to introducing an additional dev set for few-shot learning. We agree with their opinion but choose to keep a \textbf{very small} dev set mainly for feasibility consideration. Given the number of experiments in our empirical study, it is infeasible to  conduct cross-validation on every single train set for hyperparameter search.} We set three $(K_{train}, K_{dev})$ in our evaluation: (2, 1),  (5, 2) and (10, 2). We follow \citet{yang-katiyar-2020-simple} taking a greedy sampling algorithm to approximately select $K$ samples for each event type.
See Appendix~\ref{subsec:construct appendix} for details and the statistics of the sampled few-shot datasets. We inherit the original test set as $\mathcal{D}_{test}$.

\noindent{\textbf{Class-transfer setting}.} 
The few-shot datasets are curated in two sub-steps: (1) Dividing both event types and sentences in the original dataset into two disjoint parts, named \textit{source dataset} and \textit{target dataset pool}, respectively. (2) Sampling few-shot samples from the target dataset pool to construct  target dataset. The same sampling algorithm as in \textit{low-resource} setting is used. Then we have the source dataset and the sampled target dataset. See Appendix~\ref{subsec:construct appendix} for details and the statistics of the sampled few-shot datasets.

\noindent{\textbf{Evaluation Metric}} We use micro-$F1$ score as the evaluation metric. To reduce the random fluctuation, the reported values of each setting are the averaged score and sample standard deviation, of results w.r.t 10 sampled few-shot datasets. 

\subsection{Evaluated methods}
We evaluate 12 representative methods, including vanilla fine-tuning, in-context learning, 5 prompt-based and 5 prototype-based methods. These methods are detailed in Appendix~\ref{subsec:ten methods appendix}.

\noindent{\textbf{Fine-tuning}} To validate the effectiveness of few-shot methods, we fine-tune a supervised classifier for comparison as a trivial baseline.

\noindent{\textbf{In-context learning}} To validate few-shot ED tasks still not well-solved in the era of Large Language Models (LLMs), we design such baseline instructing LLMs to detect event triggers by the means of in-context learning (ICL). 

\noindent{\textbf{Prompt-based}}
(1) \textit{EEQA} (QA-based,~\citealt{du-cardie-2020-event}), (2) \textit{EETE} (NLI-based,~\citealt{lyu-etal-2021-zero}), (3) \textit{PTE} (cloze task,~\citealt{schick-schutze-2021-just}), (4) \textit{UIE} (generation,~\citealt{lu-etal-2022-unified}) and (5) \textit{DEGREE}  (generation,~\citealt{hsu-etal-2022-degree}). 

\noindent{\textbf{Prototype-based}}
(1) \textit{ProtoNet}~\cite{Snell_2017}, (2) \textit{L-TapNet-CDT}~\cite{hou-etal-2020-shot}, (3) \textit{PA-CRF}~\cite{cong-etal-2021-shot}, (4) \textit{CONTAINER}~\cite{das-etal-2022-container} and (5) \textit{FSLS}~\cite{ma-etal-2022-label}. See Table~\ref{tab:universal framework} and Figure~\ref{fig:model arch} for more details. 

\subsection{Implementation details}
We unify PLMs in each method as much as possible for a fair comparison in our empirical study. Specifically, we use \texttt{RoBERTa-base}~\cite{2019_roberta} for all prototype-based methods and three non-generation prompt-based methods. However, we keep the method's original PLM for two prompt-based methods with generation prompt, UIE (\texttt{T5-base},~\citealt{2019-T5}) and DEGREE (\texttt{BART-large},~\citealt{lewis-etal-2020-bart}). We observe their performance collapses with smaller PLMs. Regarding ICL method, we use ChatGPT (\texttt{gpt-3.5-turbo-0301}) as the language model. See more details in Appendix~\ref{subsec:implementation details}.
\section{Results: Low-resource Learning }
\label{sec:low-resource learning}

% acknowledge the limitation of our prompt design. not very diverse.
\begin{table*}
 \centering
 \caption{
 Overall results of \textit{fine-tuning} method, 10 existing few-shot ED methods, and the \textit{unified baseline} under low-resource setting. The best results are in bold face and the second best are underlined. The results are averaged over 10 repeated experiments, and sample standard deviations are in the round bracket. The standard deviations are derived from different \textbf{sampling few-shot datasets} instead of \textbf{random seeds}. Thus high standard deviation values do not mean that no significant difference among these methods.
 }
 \label{tab: overall performance}
 \setlength\tabcolsep{4.5pt}
 \small
    \begin{threeparttable}
        \begin{tabular}{@{}ll|ccc|ccc|ccc@{}} 
        \toprule
        \multicolumn{2}{c|}{\textbf{Method}} & \multicolumn{3}{c|}{\textbf{ACE05}} & \multicolumn{3}{c|}{\textbf{MAVEN}} & \multicolumn{3}{c}{\textbf{ERE}}\\
         & & {2-shot} & 5-shot & 10-shot & 2-shot & 5-shot & 10-shot & 2-shot & 5-shot & 10-shot \\
         \midrule
         \multicolumn{2}{c}{\textit{Fine-tuning}} \vline & $33.3${\tiny $(4.4)$} & $42.5${\tiny $(4.6)$} & $48.2${\tiny $(1.5)$} & $40.8${\tiny $(4.7)$}&  $52.1${\tiny $(0.7)$} & $55.7${\tiny $(0.2)$} & $32.9${\tiny $(2.1)$} & $39.8${\tiny $(2.9)$} & $43.6${\tiny $(1.7)$} \\
        \multicolumn{2}{c}{\textit{In-context Learning}} \vline &  $38.9${\tiny $(3.0)$} &   $34.3${\tiny $(1.2)$} &  $36.7${\tiny $(0.8)$} & $22.1${\tiny $(1.0)$} & $22.7${\tiny $(0.3)$} &  $23.9${\tiny $(0.7)$} & $24.2${\tiny $(3.3)$} & $26.0${\tiny $(0.7)$} & $25.5${\tiny $(1.7)$}   \\
        \midrule
          \parbox[t]{1mm}{\multirow{5}{*}{\rotatebox[origin=c]{90}{Prompt-based}}}
          & EEQA & $24.1${\tiny $(12.2)$} & $43.1${\tiny $(2.7)$} & $48.3$ {\tiny $(2.4)$} &$33.4${\tiny $(9.2)$}  &$48.1${\tiny $(0.9)$}  & $52.5${\tiny $(0.5)$}  & $13.7${\tiny $(8.6)$} & $34.4${\tiny $(1.7)$} & $39.8${\tiny $(2.4)$} \\
        &  EETE & $15.7${\tiny $(0.6)$} & $19.1${\tiny $(0.3)$} & $21.4${\tiny $(0.2)$} & $28.9${\tiny $(4.3)$} & $30.6${\tiny $(1.3)$} & $32.5${\tiny $(1.1)$} & $10.6${\tiny $(2.3)$} & $12.8${\tiny $(2.2)$} & $13.7${\tiny $(2.8)$} \\
         & PTE & $38.4${\tiny $(4.2)$} & $42.6${\tiny $(7.2)$} & $49.8${\tiny $(1.9)$} & $41.3${\tiny $(1.4)$}& $46.0${\tiny $(0.6)$}  & $49.5${\tiny $(0.6)$} & $33.4${\tiny $(2.8)$} &  $36.9${\tiny $(1.3)$} & $37.0${\tiny $(1.8)$} \\
         & UIE &$29.3${\tiny $(2.9)$} &$38.3${\tiny $(4.2)$} &$43.4${\tiny $(3.5)$} & $33.7${\tiny $(1.4)$} & $44.4${\tiny $(0.3)$} & $50.5${\tiny $(0.5)$} & $19.7${\tiny $(1.5)$} & $30.8${\tiny $(1.9)$} & $34.1${\tiny $(1.6)$}  \\
         & DEGREE & $40.0${\tiny $(2.9)$} & $45.5${\tiny $(3.2)$} & $48.5${\tiny $(2.1)$} & $43.3${\tiny $(1.0)$} & $43.4${\tiny $(5.9)$} & $45.5${\tiny $(4.3)$} & $31.3${\tiny $(3.1)$} & $36.0${\tiny $(4.6)$} & $40.7${\tiny $(2.2)$} \\
         \midrule
        \parbox[t]{1mm}{\multirow{5}{*}{\rotatebox[origin=c]{90}{Prototype-bsd}}}
        & ProtoNet & $38.3${\tiny $(5.0)$}  & $47.2${\tiny $(3.9)$} & $52.3${\tiny $(2.4)$} & $44.5${\tiny $(2.2)$}  & $51.7${\tiny $(0.6)$} & $55.4${\tiny $(0.2)$} & $31.6${\tiny $(2.7)$} & $39.7${\tiny $(2.4)$} & $44.3${\tiny $(2.3)$} \\
        & PA-CRF & $34.9${\tiny $(7.2)$} & $48.1${\tiny $(3.9)$} & $51.7${\tiny $(2.6)$} & $44.8${\tiny $(2.2)$} & $51.8${\tiny $(1.0)$} & $55.3${\tiny $(0.4)$} & $30.6${\tiny $(2.8)$} & $38.0${\tiny $(3.9)$} & $40.4${\tiny $(2.0)$} \\
        & L-TapNet-CDT & $\underline{43.2}${\tiny $(3.8)$}  & $\underline{49.8}${\tiny $(2.9)$}  & $\underline{53.5}${\tiny $(3.4)$} & $\underline{48.6}${\tiny $(1.2)$} & $\underline{53.2}${\tiny $(0.4)$} & ${56.1}${\tiny $(0.9)$} & $\underline{35.6}${\tiny $(2.6)$} &  $\underline{42.7}${\tiny $(1.7)$} & $\underline{45.1}${\tiny $(3.2)$}\\
        & CONTAINER & ${40.1}${\tiny $(3.8)$} & ${47.7}${\tiny $(3.3)$} & ${50.1}${\tiny $(1.8)$} &  $44.2${\tiny $(1.4)$} &  $50.8${\tiny $(0.9)$} &  $52.9${\tiny $(0.3)$} & $34.4${\tiny $(3.6)$} & $39.3${\tiny $(1.9)$} & $44.5${\tiny $(2.3)$}\\
        & FSLS & $39.2${\tiny $(3.4)$} & $47.5${\tiny $(3.2)$}  & $51.9${\tiny $(1.7)$} & $46.7${\tiny $(1.2)$} & $51.5${\tiny $(0.5)$} & $\underline{56.2}${\tiny $(0.2)$} & $34.5${\tiny $(3.1)$} & $39.8${\tiny $(2.5)$} & $44.0${\tiny $(2.0)$} \\
         \midrule 
        \multicolumn{2}{c}{Unified Baseline} \vline & $\textbf{46.0}${\tiny $(4.6)$} & $\textbf{54.4}${\tiny $(2.6)$}  & $\textbf{56.7}${\tiny $(1.5)$} & $\textbf{49.5}${\tiny $(1.7)$}  & $\textbf{54.7}${\tiny $(0.8)$} & $\textbf{57.8}${\tiny $(1.2)$} & $\textbf{38.8}${\tiny $(2.4)$} & $\textbf{45.5}${\tiny $(2.8)$} & $\textbf{48.4}${\tiny $(2.6)$} \\
        \bottomrule
        \end{tabular}
    \end{threeparttable}
    \label{Ablation study}
\end{table*}

\subsection{Overall comparison}
We first overview the results of the 12 methods under the low-resource setting in Table~\ref{tab: overall performance}. 

\noindent{\textbf{Fine-tuning}.} Despite its simpleness, fine-tuning achieves acceptable performance. In particular, it is even comparable to the strongest existing methods on MAVEN dataset, only being $1.1\%$ and $0.5\%$ less under 5-shot and 10-shot settings. One possible reason that fine-tuning is good on MAVEN is that MAVEN has 168 event types, much larger than others. When the absolute number of samples is relatively large, PLMs might capture implicit interactions among different event types, even though the samples per event type are limited. When the sample number is scarce, however, fine-tuning is much poorer than existing competitive methods (see ACE05). Thus, we validate the necessity and progress of existing few-shot methods. 

\noindent{\textbf{In-context learning}.} We find the performance of ICL-based methods lags far behind that of tuning-required methods, though the backbone of ICL approach (ChatGPT) is much larger than other PLMs (<1B). A series of recent work~\cite{ma2023large, 2023_chatgpt_ee, zhan2023glen} observe the similar results as ours~\footnote{We refer readers to \citet{ma2023large} for a more detailed discussion on why ICL approaches stumble across few-shot ED tasks.}. Thus we validate few-shot ED tasks could not be solved smoothly by cutting-edge LLMs and deserves further exploration. 

\noindent{\textbf{Prompt-based methods}.} 
Prompt-based methods deliver much poorer results than expected, even compared to fine-tuning, especially when the sample number is extremely scarce. It shows designing effective prompts for ED tasks with very limited annotations is still challenging or even impossible. We speculate it is due to the natural gap between ED tasks and pre-training tasks in PLMs.

Among prompt-based methods, PTE and DEGREE achieve relatively robust performance under all settings. DEGREE is advantageous when the sample size is small, but it cannot well handle a dataset with many event types like MAVEN. When sample sizes are relatively large, EEQA shows competitive performance as well.

\subsection{Prototype-based methods} 
\label{subsec: analyze on prototype-based methods, low-resource settings}
Since prototype-based methods have overall better results, we zoom into the design elements to search for effective choices based on the unified view.

\begin{figure}[htbp!]
\centering
    \includegraphics[width=\linewidth]{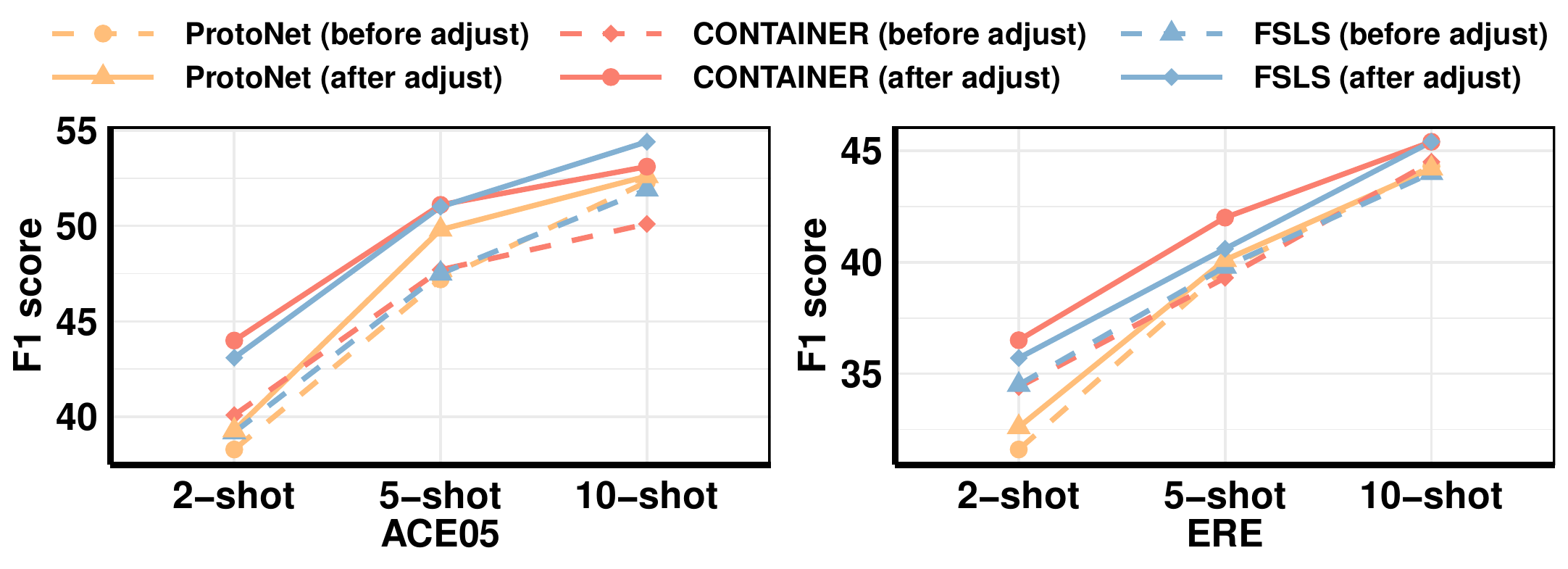}
    \caption{Results of existing methods \textit{before} (dashed lines) and \textit{after} (solid lines) adjustment that substitute their transfer and distance functions to appropriate ones. See full results in Table~\ref{tab:d and f}.}
    \label{fig:existing methods after adjustment}
\end{figure}

\noindent{\textbf{Transfer function, Distance function, and CRF.}}
We compare combinations of transfer and distance functions and four variants of CRF modules in Appendices~\ref{subsec:appendix d and f} and~\ref{subsec:appendix CRF module}. We make two findings: (1) A scaled coefficient in the distance function achieves better performance with the normalization transfer function. 
(2) There is no significant difference between models with or without CRF modules. Based on these findings, we observe a significant improvement in five existing methods by simply substituting their $d$ and $f$ for more appropriate choices, see Figure~\ref{fig:existing methods after adjustment} and Appendix~\ref{subsec:appendix d and f}. We would use these new transfer and distance functions in further analysis and discussion.

\begin{figure}[htbp!]
    \centering
    \includegraphics[width=\linewidth]{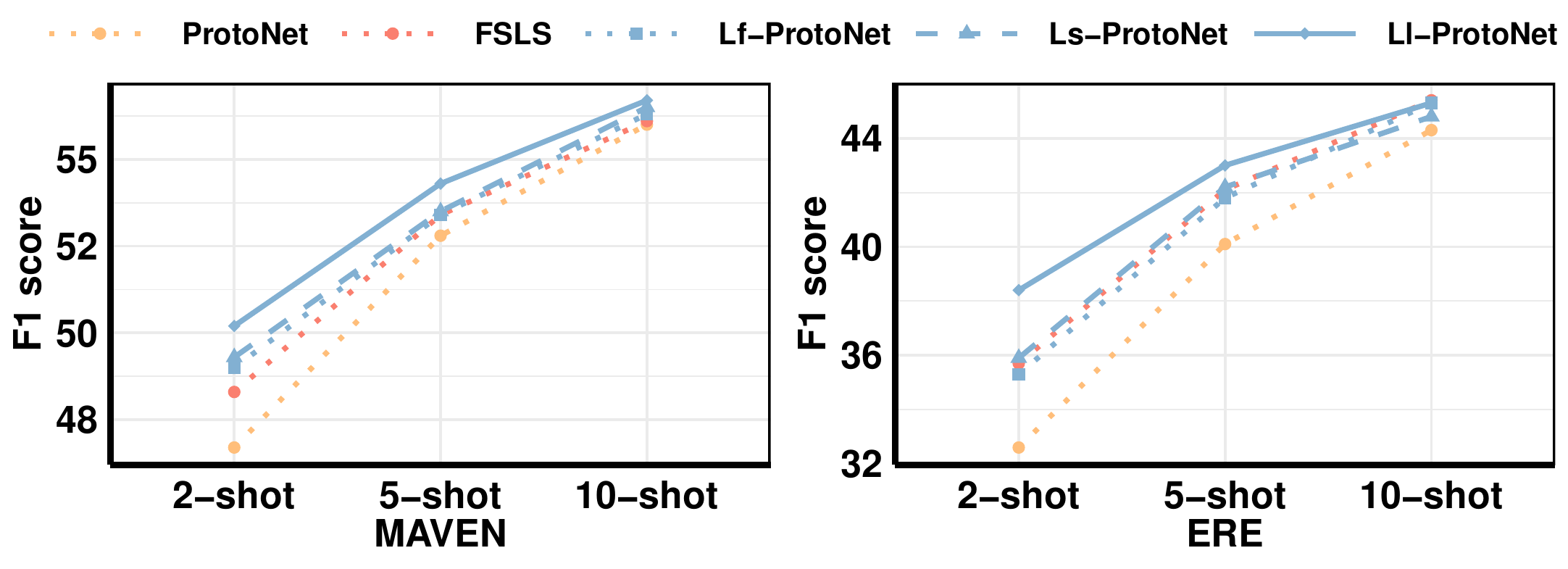}
    \caption{Results of three approaches aggregating label semantics and event mentions on MAVEN and ERE few-shot datasets. \textbf{Lf}: feature-level. \textbf{Ls}: score-level. \textbf{Ll}: loss-level. See full results in Table~\ref{tab:aggregation form}.}
    \label{fig:aggregation form}
\end{figure}

\noindent{\textbf{Prototype Source.}}
We explore whether label semantic and event mentions are complementary prototype sources, i.e., whether utilizing both achieves better performance than either one. 
We choose ProtoNet and FSLS as base models which contain only a single kind of prototype source (mentions or labels). Then we combine the two models using three aggregating forms mentioned in Section~\ref{sec:unified view} and show their results in Figure~\ref{fig:aggregation form}. Observe that: (1) leveraging label semantics and mentions as prototype sources simultaneously improve the performance under almost all settings, and (2) merging the two kinds of sources at loss level is the best choice among three aggregation alternatives.

\noindent{\textbf{Contrastive or Prototypical Learning}.} Next, we investigate the effectiveness of contrastive learning (CL, see CONTAINER) and prototypical learning (PL, see ProtoNet and its variants) for event mentions. We compare three label-enhanced (since we have validated the benefits of label semantics) methods aggregating event mentions with different approaches. (1) \textit{Ll-ProtoNet}: the strongest method utilizing PL in last part. (2) \textit{Ll-CONTAINER}: the method utilizing in-batch CL as CONTAINER does. (3) \textit{Ll-MoCo}: the method utilizing CL with MoCo setting~\cite{2020_MoCo}. The in-batch CL and MoCo CL are detailed in Appendix~\ref{subsec:appendix CL}.

Figure~\ref{fig:in-batch or moco} suggests CL-based methods outperform Ll-ProtoNet. There are two possible reasons: (1) CL has higher sample efficiency since every two samples interact during training. PL, however, further splits samples into support and query set during training; samples within the same set are not interacted with each other.  (2) CL adopts score-level aggregation while PL adopts feature-level aggregation. We find the former also slightly outperforms the latter in Figure~\ref{fig:aggregation form}. We also observe that MoCo CL usually has a better performance than in-batch CL when there exists complicated event types (see MAVEN), or when the sample number is relatively large (see ACE 10-shot). We provide a more detailed explanation in Appendix~\ref{subsec:appendix CL}.

\begin{figure}[htbp!]
    \centerline{\includegraphics[width=\linewidth]{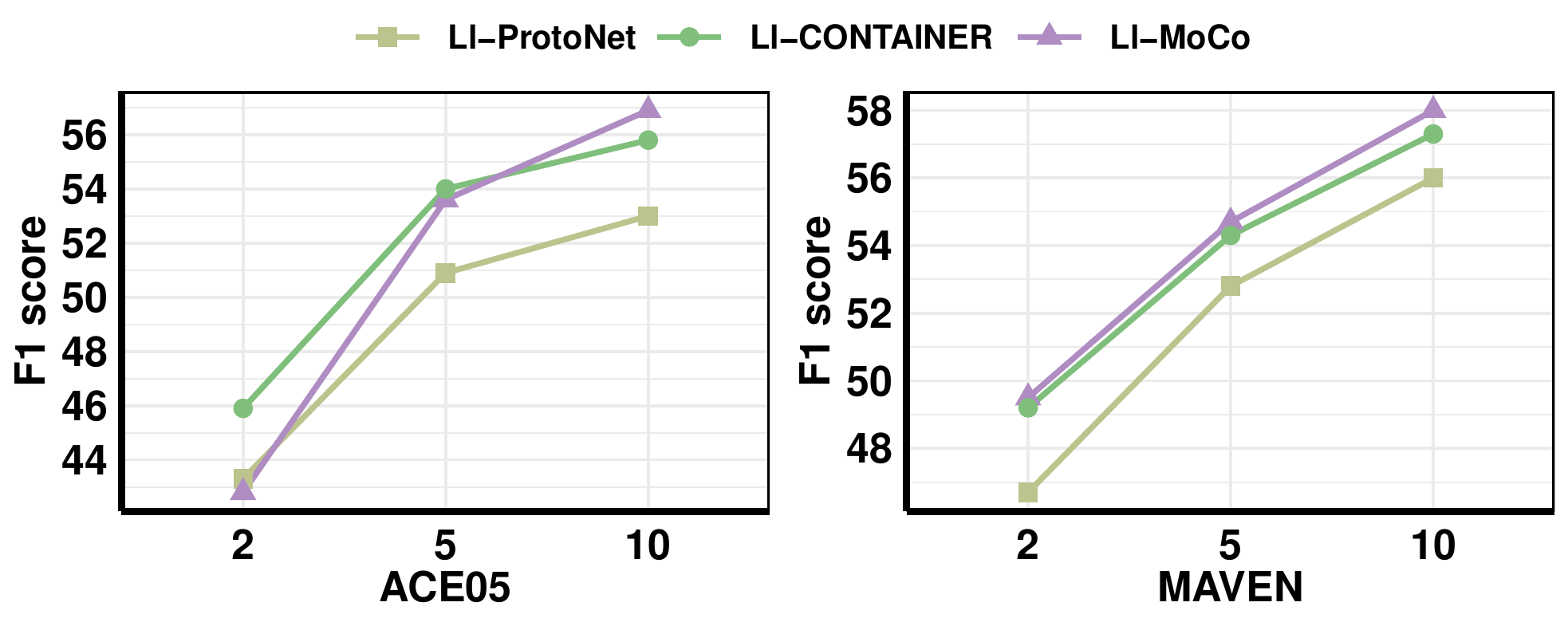}}
    \caption{Results of (label-enhanced) PL and CL methods on ACE05 and MAVEN few-shot datasets. See full results on three datasets in Table~\ref{tab:in-batch or moco}. }
    \label{fig:in-batch or moco}
\end{figure}

\subsection{The unified baseline}
\label{subsec:our baseline}
Here is a summary of the findings: (1) Scaled euclidean or cosine similarity as distance measure with normalized transfer benefits existing methods. (2) CRF modules show no improvement in performance. (3) Label semantic and event mentions are complementary prototype sources, and aggregating them at loss-level is the best choice. (4) As for the branch of event mentions, CL is more advantageous than PL for few-shot ED tasks. (5) MoCo CL performs better when there are a good number of sentences, otherwise in-batch CL is better.

Based on these findings, we develop a simple but effective \textit{unified baseline} as follows. We utilize \underline{both label semantic and event mentions} as prototype sources and aggregate two types of sources at \underline{loss-level}. Specifically, we assign two branches with their own losses for label semantic and event mentions respectively. Both two branches adopt \underline{scaled cosine similarity} $d_\tau(u, v) = -\frac{u^Tv}{\tau}$ as distance measure and \underline{normalization} $f(h) = h/\|h\|_2$ as transfer function. We do not add CRF modules. 

For label semantic branch, we follow FSLS and set the embeddings of event name as prototypes. Here $h_x$ and $h_{e_y}$ represent the PLM representation of event mention $x$ and label name $e_y$, respectively.

\begin{equation}
\begin{aligned}
\nonumber
    e_y &= \text{Event\_name}(y) \\
    \text{logits}^{(l)}(y|x) &= -d_\tau(f(h_x), f(h_{e_y}))
\end{aligned}
\end{equation}

For event mention branch, we adopt CL which aggregates prototype sources (event mentions) at score-level. If the total sentence number in train set is smaller than 128, we take in-batch CL (CONTAINER) strategy as below:
\begin{equation}
\begin{aligned}
\nonumber
     \text{logits}^{(m)}(y|x) = \sum_{x' \in \mathcal{S}_{y}(x)} \frac{-d(f(h_x), f(h_{x'}))}{|\mathcal{S}_{y}(x)|}
\end{aligned}
\end{equation}

$\mathcal{S}_y(x) = \{x'|(x', y') \in D, y'=y, x' \neq x \}$ is the set of all other mentions with the same label. If the total sentence number in train set is larger than 128, we instead take MoCo CL maintaining a queue for $\mathcal{S}_y(x)$ and a momentum encoder.

We then calculate the losses of these two branches and merge them for joint optimization:
\begin{equation}
\begin{aligned}
\nonumber
    p^{(l/m)}(y|x) &= \text{Softmax}_y [\text{logits}^{(l/m)}(y|x)] \\
    L^{(l/m)}(y|x) &= - \sum_{(x,y)} y \text{log}(p^{(l/m)}(y|x)) \\
    L &= L^{(l)} + L^{(m)}
\end{aligned}
\end{equation}

The diagram of the \textit{unified baseline} is illustrated in Figure~\ref{fig:model arch}(f) and its performance is shown in Table~\ref{tab: overall performance}. Clearly, \textit{unified baseline} outperforms all existing methods significantly, 2.7$\%$ $F$1 gains on average, under all low-resource settings.
\section{Results: Class-transfer Learning}
\label{sec:class-transfer learning}
In this section, we evaluate existing methods and the \textit{unified baseline} under class-transfer setting. Here we do not consider in-context learning because previous expetiments show it still lags far from both prompt- and prototype-based methods.

\subsection{Prompt-based methods}
\label{subsec:class-transfer-prompt}
\begin{figure}[!htbp]
\centering
    \subfigure[ACE05]{
    \includegraphics[width=\linewidth]{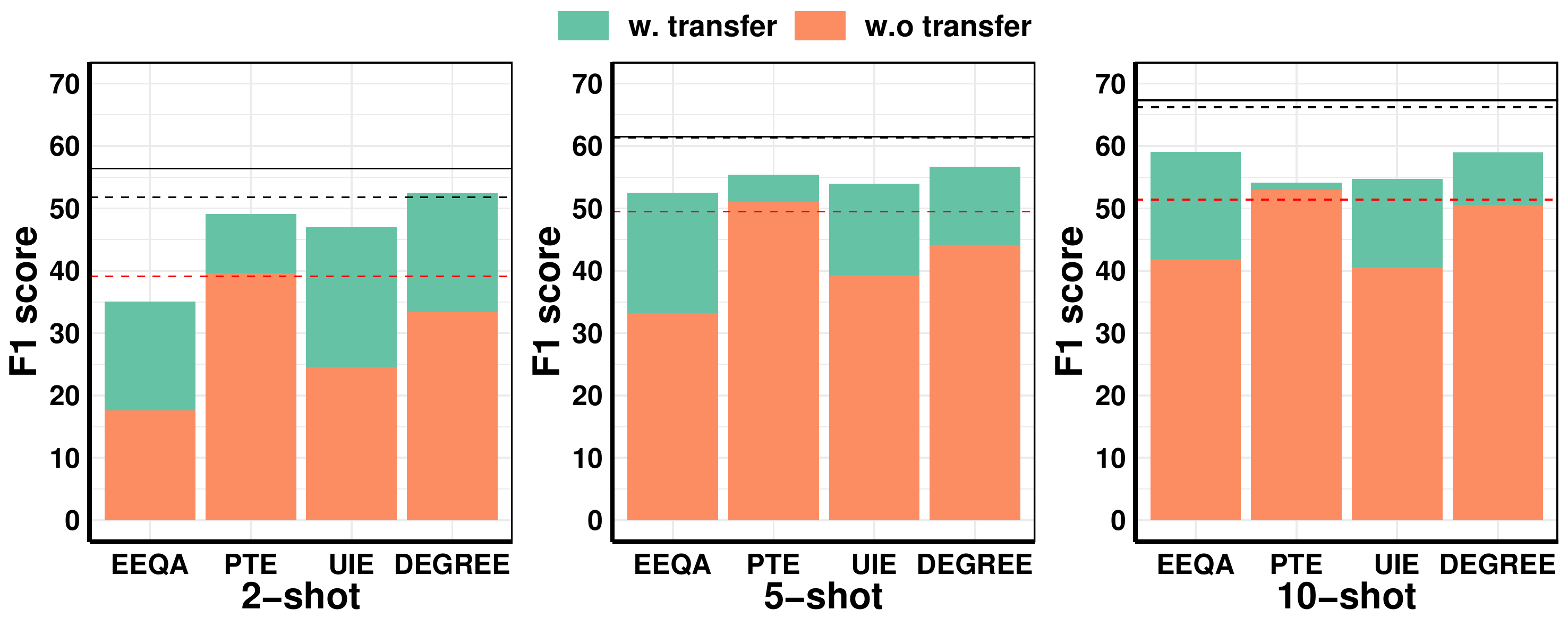}
    }
    
    \subfigure[MAVEN]{
    \includegraphics[width=\linewidth]{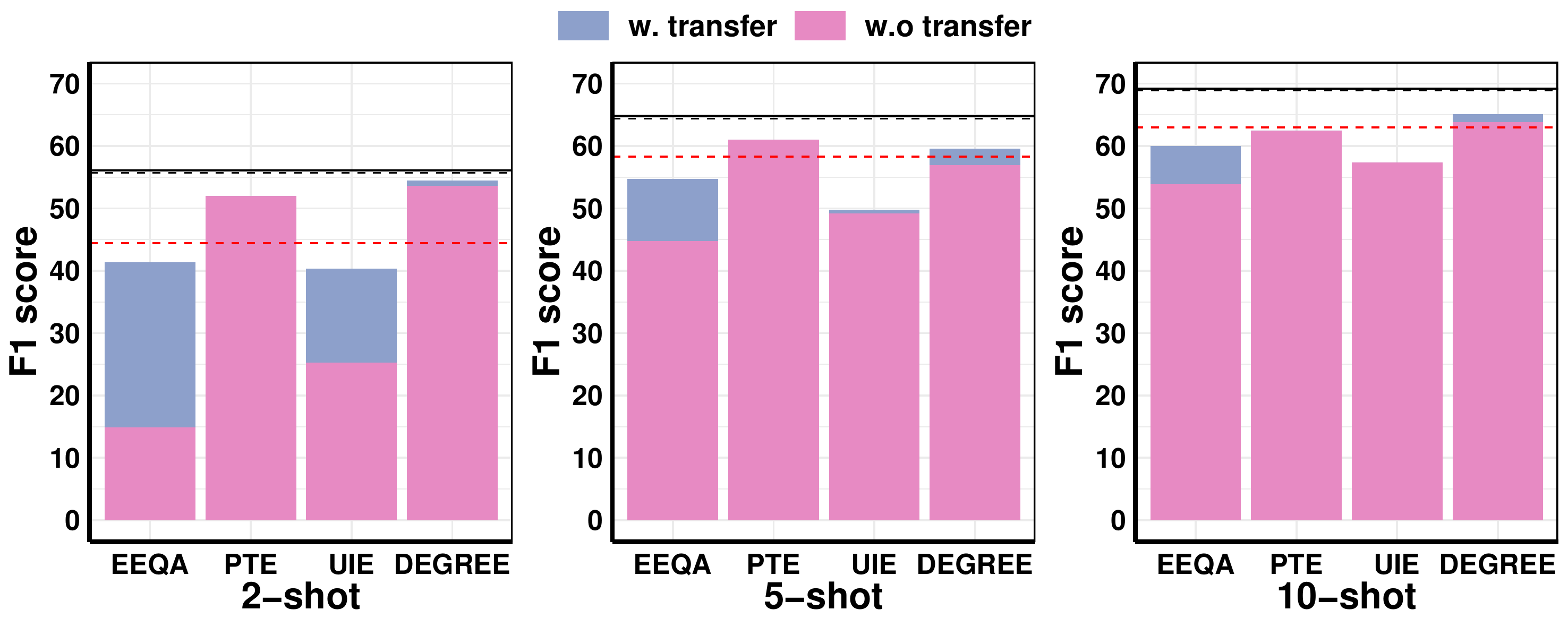}
    }
\caption{Class-transfer results of prompt-based methods. We plot \textit{fine-tuning}  (red dash lines), best and second best prototype-based methods (black solid/dash lines) for comparison. See full results in Table~\ref{tab:class-transfer-prompt}.}
\label{fig:domain-transfer-prompt}
\end{figure}

We first focus on 4 existing prompt-based methods and explore whether they could smoothly transfer event knowledge from a preexisting (source) schema to a new (target) schema. We show results in Figure~\ref{fig:domain-transfer-prompt} and Appendix~\ref{subsec: appendix class transfer prompt}. The findings are summarized as follows. (1) The transfer of knowledge from source event types to target event types facilitates the model prediction under most scenarios. It verifies that an appropriate prompt usually benefits inducing the knowledge learned in PLMs. (2) However, such improvement gradually fades with the increase of sample number from either source or target schema. For example, the 5-shot v.s 10-shot performance for PTE and UIE are highly comparable. We speculate these prompts act more like a catalyst: they mainly teach model how to induce knowledge from PLMs themselves rather than learn new knowledge from samples. Thus the performance is at a standstill once the sample number exceeds some threshold. (3) Overall, the performance of prompt-based methods remains inferior to prototype-based methods in class-transfer setting (see black lines in Figure~\ref{fig:domain-transfer-prompt}). Since similar results are observed in low-resource settings as well, we conclude that prototype-based methods are better few-shot ED task solver.

\subsection{Prototype-based methods}
\label{subsec:class-transfer-prototype}

\begin{figure*}[!htbp]
\centering
    \includegraphics[width=\linewidth]{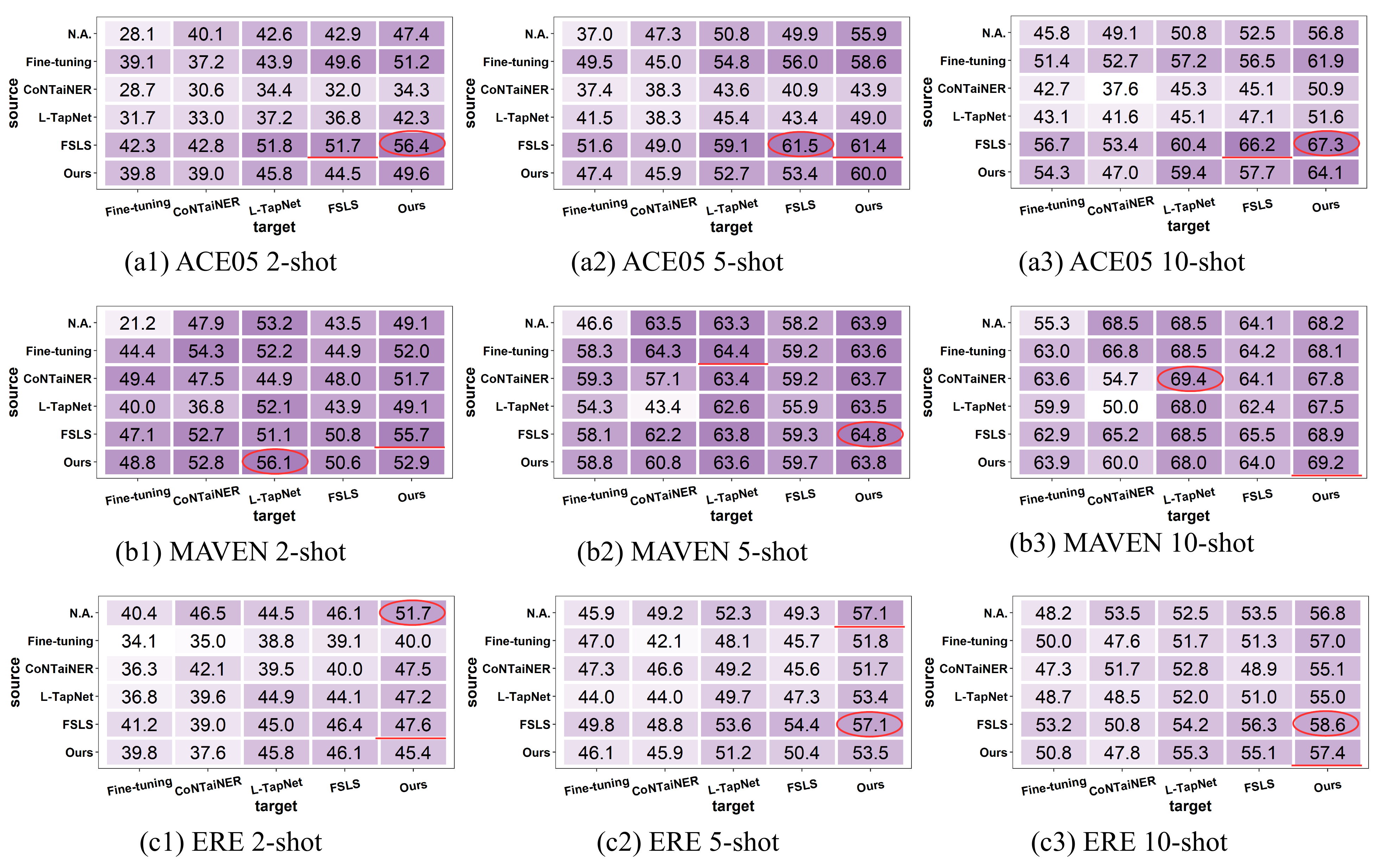}
    \caption{
    Class-transfer results of \textit{fine-tuning} methods and four prototype-based methods on three datasets. For each matrix, row and column represent the source and target models,  respectively. For example, the value in top-left corners of every matrix means the performance when directly finetuning a model in target dataset (source: N.A. / target: Fine-tuning).
    Each value is the results averaged over 10 repeated experiments. See full results in Table~\ref{tab:class-transfer-prototype}.
    }
    \label{fig:domain-transfer-prototype}
\end{figure*}

We further explore the transfer ability of existing prototype-based methods and \textit{unified baseline}\footnote{Transfer and distance functions in all methods are substituted to appropriate ones and CRF modules are removed.}. Thanks to the unified view, we conduct a more thorough experiment that enumerates all possible combinations of models used in the source and target domain, to assess if the generalization ability affects transferability. That is, the parameters in PLMs will be shared from source to target model. We show results in Figure~\ref{fig:domain-transfer-prototype} and Appendix~\ref{subsec: appendix class transfer prototype}.

\noindent{\textit{1. Is transfer learning effective for prototype-based methods?}} It depends on the dataset (compare the first row with other rows in each column). For ACE05 and MAVEN datasets, the overall answer is yes. Contrary to our expectation, transfer learning affects most target models on ERE dataset negatively, especially for 2- and 5-shot settings.

\noindent{\textit{2. Do prototype-based methods perform better than simple fine-tuning?}} It depends on whether \textit{fine-tuning} the source or target model. When \textit{fine-tuning} a source model (row 2), it sometimes achieves comparable even better performance than the prototype-based methods (last 4 rows). When \textit{fine-tuning} a target model (column 1), however, the performance drops significantly. Thus, we speculate that powerful prototype-based methods are more necessary in target domain than source domain.

\noindent{\textit{3. Is the choice of prototype-based methods important?}} Yes. When we select inappropriate prototype-based methods, they could achieve worse performance than simple fine-tuning and sometimes even worse than models without class transfer. For example, CONTAINER and L-TapNet are inappropriate source model for ACE05 dataset.

\noindent{\textit{4. Do the same source and target models benefit the event-related knowledge transfer?}} No. 
The figures show the best model combinations often deviate from the diagonals. It indicates that different source and target models sometimes achieve better results.

\noindent{\textit{5. Is there a source-target combination performing well on all settings?}} Strictly speaking, the answer is No. Nevertheless, we find that adopting FSLS as the source model and our \textit{unified baseline} as the target model is more likely to achieve competitive (best or second best) performance among all alternatives. It indicates that (1) the quality of different combinations show kinds of \textbf{tendency} though no consistent conclusion could be drawn. (2) a model with moderate inductive bias (like FSLS) might be better for the source dataset with abundant samples. Then our \textit{unified baseline} could play a role during the target stage with limited samples.
\section{Conclusion}

We have conducted a comprehensive empirical study comparing 12 representative methods under unified \textit{low-resource} and \textit{class-transfer} settings. For systematic analysis, we proposed a unified framework of promising prototype-based methods. Based on it, we presented a simple and effective \textit{baseline} that outperforms all existing methods significantly under \textit{low-resource} setting, and is an ideal choice as the target model under \textit{class-transfer} setting. In the future, we aim to explore how to leverage unlabeled corpus for few-shot ED tasks, such as data augmentation, weakly-supervised learning, and self-training.
\section*{Acknowlegement}
This study is supported under the RIE2020 Industry Alignment Fund – Industry Collaboration Projects (IAF-ICP) Funding Initiative, the Singapore Ministry of Education (MOE) Academic Research Fund (AcRF) Tier 1 grant, as well as cash and in-kind contribution from the industry partner(s).
\section*{Limitations}
We compare 12 representative methods, present a \textit{unified view} on existing prototype-based methods, and propose a competitive \textit{unified baseline} by combining the advantageous modules of these methods. We test all methods, including the unified baseline, on three commonly-used English datasets using various experimental settings and achieve consistent results. However we acknowledge the potential disproportionality of our experiments in terms of language, domain, schema type and data scarcity extent. Therefore, for future work, we aim to conduct our empirical studies on more diverse event-detection (ED) datasets.

We are fortunate to witness the rapid development of Large Language Models (LLMs~\citealt{2020_gpt3, 2022_instructgpt, 2022_flant5}) in recent times. In our work, we set in-context learning as a baseline and evaluate the performance of LLMs on few-shot ED tasks. We find current LLMs still face challenges in dealing with Information Extraction (IE) tasks that require structured outputs~\cite{qin2023_benchmark, 2023_synIE}. However, we acknowledge the ICL approach adopted here is relatively simple. We do not work hard to find the optimal prompt format, demonstration selection strategy, etc., to reach the upper bounds of LLMs' performance. We view how to leverage the power of LLMs on ED tasks as an open problem and leave it for future work.

In this work, we focus more on the model aspect of few-shot ED tasks rather than data aspect. In other words, we assume having and only having access to a small set of labeled instances. In the future, we plan to explore how to utilize annotation guidelines, unlabeled corpus and external structured knowledge to improve few-shot ED tasks.

\bibliography{anthology, custom}
\bibliographystyle{acl_natbib}

\appendix
\section{Related Work}
\label{sec:related work}

\subsection{Taxonomy of task settings}
\label{subsec:four settings}
\begin{figure*}
    \centering
    \includegraphics[width=\linewidth]{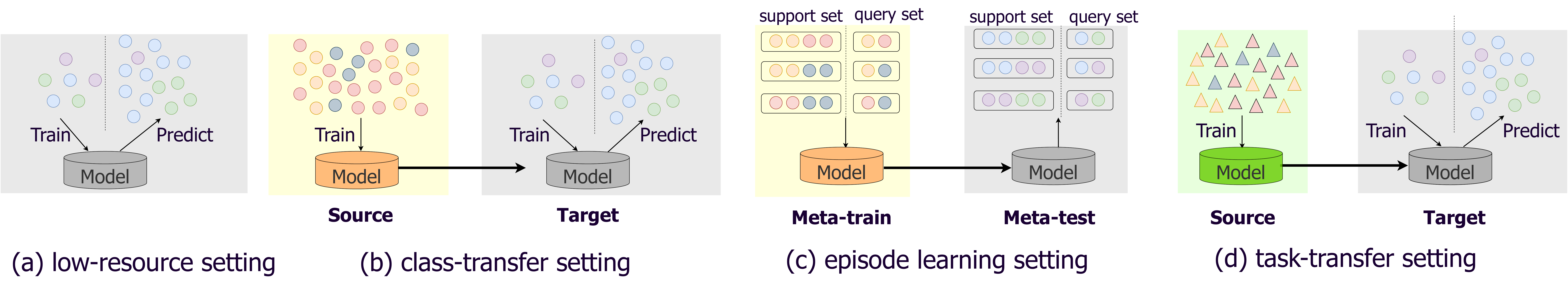}
    \caption{Four few-shot settings summarized from previous work. Different colors represent different event types. Different shapes represent samples with different tasks.}
    \label{fig:four definitions}
\end{figure*}

Various solutions have been proposed to improve the \textit{generalization} and \textit{transfer} abilities of few-shot ED methods. There exists a bottleneck: the models adopt very different tasks and experimental settings. We categorize existing task settings to four cases as shown in Figure~\ref{fig:four definitions}: \textit{low-resource} (LR), \textit{class transfer} (CL), \textit{episode learning} (EL), and \textit{task transfer} (TT) settings. LR is used to evaluate the \textit{generalization} ability, learning rapidly with only few examples in target domain. The other settings (CL, EL, and TT) evaluate the \textit{transfer} ability, adapting a model trained with a preexisting schema with abundant samples, to a new (target) schema with only few examples. Based on the pros and cons presented here, we adopt the \textit{low-resource}  and \textit{class transfer} settings in our empirical study. 

\noindent{\textbf{1. Low-resource setting}} assesses the generalization ability of models by (1) utilizing only few samples during training, (2) evaluating on the real and rich test dataset.  Conventionally, the few-shot $|\mathcal{D}_{train}|$ and $|\mathcal{D}_{dev}|$ are downsampled from a full dataset by two main strategies: (1) \textit{$K$-shot sampling} which picks out $K$ samples for each event type, or (2) \textit{ratio sampling} which picks out partial sentences with a fixed ratio. We view both sampling strategies as reasonable and adopt $K$-shot sampling in this work.

The surging development of PLMs makes training with only few (or even zero) examples possible, and achieves acceptable performance~\cite{devlin-etal-2019-bert, 2019-T5, 2020-gpt3}. Accordingly, a series of prompt-based methods~\cite{du-cardie-2020-event, liu-etal-2020-event, 2020_Feng, 2021_Paolini, lu-etal-2021-text2event, deng-etal-2021-ontoed, hsu-etal-2022-degree, Li_2022_PILED} adopt such setting to train and evaluate their models.

\noindent{\textbf{2. Class transfer setting}} assesses the \textit{transferability} of a model by providing abundant samples in the source (preexisting) schema and scarce samples in target (new) schema. It trains a classifier in source schema and then transfers such classifier to the target schema with only few examples.

Such setting has been applied since an early stage~\cite{bronstein-etal-2015-seed, peng-etal-2016-event, zhang-etal-2021-zero}, and is often used together with low-resource setting to additionally evaluate transferability of the models~\cite{2021_Paolini, lu-etal-2021-text2event, hsu-etal-2022-degree}.

\noindent{\textbf{3. Episode learning setting}} is a classical few-shot setting. It has two phases, \textit{meta-training} and \textit{meta-testing}, each of which consists of multiple episodes. Each episode is a few-shot problem with its own train (support) and test (query) sets and event-type classes. Since the sets in each episode are sampled uniformly having $K$ different classes and each class having $N$ instances, episode learning is also known as \textbf{$N$-way-$K$-shot} classification. 

Many existing few-shot ED methods adopt this setting~\cite{Lai_2020, lai-etal-2020-extensively, Deng_2020, cong-etal-2021-shot, lai-etal-2021-learning, chen-etal-2021-honey, zhang-etal-2022-hcl, zhao-etal-2022-knowledge}. However, we argue that episode learning assumes an unrealistic scenario. First, during the meta-training stage, a large number of episodes is needed, for example, 20,000 in \citet{cong-etal-2021-shot}. Though the label sets of meta-training and meta-testing stages are disjoint, class transfer setting is more reasonable when there are many samples in another schema. Second, tasks with episode learning are evaluated by the performance on samples of the test (query) set in the meta-testing phase. The test sets are sampled uniformly, leading to a significant discrepancy with the true data distribution in many NLP tasks. The absence of sentences without any events further leads to distribution distortion. Further, each episode contains samples with only $K$ different classes, where $K$ is usually much smaller than the event types in the target schema. All these factors may lead to an overestimation on the ability of few-shot learning systems. For above reasons, we do not consider this setting in our experiments.

\noindent{\textbf{4. Task transfer setting}} is very similar to class transfer. The main difference is that  it relaxes the constraint in source phase, from the same task with different schema to different tasks.\footnote{Generally speaking, all methods using PLMs belong to this setting in which the source task is exactly the pre-training task of PLMs, masked- or next-word prediction. In this work, we limit the discussion of task transfer to which the source task is another downstream task rather than the general pre-training task in PLMs.} The development of this setting also heavily relies on the success of PLMs. \citet{liu-etal-2020-event}, \citet{2020_Feng} and \citet{lyu-etal-2021-zero} leverage model pre-trained with SQuAD 2.0 (QA dataset, ~\citealt{rajpurkar-etal-2018-know}) or MNLI (NLI dataset, ~\citealt{williams-etal-2018-broad}) to improve the performance of zero-/few-shot ED models. \citet{2021_Paolini} and \citet{lu-etal-2022-unified} recently construct unified generation frameworks on multiple IE tasks. Their experiments also reveal that pre-training on these tasks benefits few-shot ED. Though task transfer setting is reasonable and promising, we do not include this setting out of its extreme diversity and complexity. That is, there are (1) too many candidate tasks as pre-training tasks, and (2) too many optional datasets for each pre-training task. Thus it is almost infeasible  to conduct a comprehensive empirical study on task transfer setting.

\subsection{Taxonomy of methods}
\label{subsec:method appendix}
We categorize existing methods to two main classes, \textbf{prompt-based} methods and \textbf{prototype-based} methods, and list them in Table~\ref{tab:method comparison}. Here we give a detailed introduction of existing methods. Note that in our empirical study, we also include some methods which are originally developed for similar few-shot tasks but can be easily adapted to ED.  We leave a special subsection for them.

\noindent{\textbf{Few-shot ED methods.}} Due to the prohibitively cost for labeling amounts of event mentions, few-shot ED is a long-standing topic in event-related research community. The proposed solutions are mainly in two branches. The first branch,  \textit{prototype-based}~\footnote{Different from other sections, here we adopt a chronological order and firstly introduce prototype-based methods.} methods, is a classical approach on few-shot learning. It defines a single or multiple \textit{prototypes} for each event type representing the label-wise properties. It then learns the embedding representation of each sample via shortening the distance from its corresponding prototypes given a distance/similarity metric. \citet{bronstein-etal-2015-seed} and \citet{peng-etal-2016-event} leverage the seed instances in annotation guideline and mine the lexical/semantic features of trigger words to obtain the prototypes. \citet{zhang-etal-2021-zero} inherit such paradigm and define prototypes as the average contextualized embeddings of the related trigger words weakly labeled in external corpus. With the help AMR Parsing, \citet{huang-etal-2018-zero} additionally consider the graph structures of preexisting schema as prototypes, and encode AMR graph representation of each event mention as representations. \citet{Deng_2020} introduces Dynamic Memory Network (DMN), while \citet{Lai_2020} and \citet{lai-etal-2021-learning} introduce two different auxiliary losses improving intra-/inter-consistency of different episodes to facilitate their prototype representations. \citet{deng-etal-2021-ontoed} further consider the relations among events to constrain the prototypes and benefit both rare and new events. \citet{cong-etal-2021-shot} amortize CRF module by modeling the transition probabilities of different event types with their prototypes. \citet{chen-etal-2021-honey} leverage causal inference and intervene on context via backdoor adjustment during training to reduce overfitting of trigger words for more robust prototypes. Recently, \citet{zhang-etal-2022-efficient-zero} and \citet{zhang-etal-2022-hcl} introduce contrastive learning into few-shot ED task and their proposed methods actually could be viewed as \textit{generalized} prototype-based methods with \textit{multiple} prototypes rather than one.

The other  branch, \textit{prompting methods}, is made possible with the surge of development in PLMs. Given a specific task, prompting methods map the task format to a new format with which the PLMs are more familiar, such as masked word prediction~\cite{schick-schutze-2021-just} and sequence generation~\cite{2019-T5, 2020-gpt3}. Such format conversion narrows down the gaps between pre-training tasks and downstream tasks, which is beneficial for inducing learned knowledge from PLMs with limited annotations. As for event detection (and many other IE tasks), however, it is not trivial to design a smooth format conversion. One simple idea is leveraging one single template to prompt both event types and their triggers simultaneously~\cite{2021_Paolini, lu-etal-2021-text2event}. However, such prompting methods show performance far from satisfactory, especially when they are not enhanced by two-stage pre-training and redundant hinting prefix~\cite{lu-etal-2022-unified}.
Another natural idea is enumerating all legal spans and querying the PLMs whether each span belongs to any class, or vice versa~\cite{hsu-etal-2022-degree}. A major limitation here is the prohibitively time complexity, particularly when there are many event types.
Combining the merits of \textit{prompting methods} and conventional \textit{fine-tuning methods} is another solution. \citet{du-cardie-2020-event} and \citet{liu-etal-2020-event} use QA/MRC format to prompt the location of trigger words, while still predicting their event types via an additional linear head. \citet{lyu-etal-2021-zero} first segment one sentence into several clauses and view the predicates of clauses as trigger candidates. 
Then they leverage NLI format to query the event types of these candidates. Recently, \citet{Li_2022_PILED} propose a strategy combining Pattern-Exploiting Training (PET,~\citealt{schick-schutze-2021-just}) and CRF module. Initially, they conduct sentence-level event detection determining whether one sentence contains any event types or not. For each identified event type, they further use a linear chain CRF to locate the trigger word.

\noindent{\textbf{Few-shot NER/ST methods.}} There are several models which are originally designed for similar tasks like Named Entity Recognition (NER) and Slot Tagging (ST) but could be applied to ED task.

Similar to ED methods, one classical paradigm in NER is utilizing ProtoNet~\cite{Snell_2017} and its variants to learn \textit{one} representative prototypes for each class type with only few examples. \citet{2019_Fritzler} firstly combine ProtoNet and CRF module to solve NER tasks. \citet{hou-etal-2020-shot} propose L-TapNet-CDT, which enhances TapNet~\cite{2019_tapnet}, a variant of ProtoNet, with textual label names and achieves great performance among several ST tasks. Both methods construct prototypes by computing the average embeddings of several sampled examples (support set). \citet{yang-katiyar-2020-simple} propose a simpler algorithm, leveraging supervised classifier learned in preexisting schema as feature extractor and adopting nearest neighbors classification during inference, and show competitive performance in class transfer setting for few-shot NER task. \citet{das-etal-2022-container} introduce contrastive learning into few-shot NER task. \citet{ma-etal-2022-label} recently developed a simple but effective method on few-shot NER by constructing prototypes only with their labels.

\section{Datasets and Models}
\label{sec:datasets and models}

We curate few-shot datasets used in this emprical study from three full and commonly-used datasets: ACE05~\cite{doddington-etal-2004-automatic}, MAVEN~\cite{wang-etal-2020-maven} and ERE~\cite{song-etal-2015-light}.

\subsection{Full dataset}
\label{subsec:full dataset}

ACE05 is a joint information extraction dataset, with 
annotations of entities, relations, and events. We only use its event annotation for ED task. It contains 599
English documents and 33 event types in total. We split documents in ACE05  following previous work~\cite{li-etal-2013-joint} to construct train and test dataset respectively. MAVEN is a newly-built large-scale ED dataset with 4480 documents and 168 event types. We use the official split for MAVEN dataset. ERE is another joint information extraction dataset having a similar scale as ACE05 (458 documents, 38 event types). We follow the preprocessing procedure in \citet{lin-etal-2020-joint}. Table~\ref{tab:full dataset} reports detailed statistics of the three datasets.

ED could be viewed as either a span classification or a sequence labeling task. In our work, we adopt span classification paradigm for MAVEN dataset since it provides official spans for candidate triggers (including negative samples). For the other two datasets, we follow sequence labeling paradigm to predict the event type word by word.

\begin{table}
\small
\centering
\caption{Statistics of three full ED datasets.}
    \label{tab:full dataset}
    \begin{tabular}{ll|rrr}
    \toprule
    \multicolumn{2}{c|}{\textbf{Dataset}} & \textbf{ACE05} & \textbf{MAVEN} & \textbf{ERE}\\
    \midrule
    \multicolumn{2}{c|}{\textbf{\#Event type}} & 33  & 168 & 38 \\
    \midrule
    \multirow{2}{*}{\textbf{\#Sents}} & Train & 14,024 & 32,360 & 14,736 \\
    & Test & 728 & 8,035 & 1,163 \\
    \midrule
    \multirow{2}{*}{\textbf{\#Mentions}} & Train & 5,349 & 77,993 & 6,208 \\
    & Test & 424 & 18,904 & 551 \\
    \bottomrule
    \end{tabular}
\end{table}

\subsection{Dataset construction}
\label{subsec:construct appendix}
This section introduces how we construct few-shot datasets from the three full ED datasets.

\noindent{\textbf{Low-resource setting.}} We downsample sentences from original full training dataset to construct $\mathcal{D}_{train}$ and $\mathcal{D}_{dev}$, and inherit the original test set as the unified $\mathcal{D}_{test}$. For $\mathcal{D}_{train}$ and $\mathcal{D}_{dev}$, we adopt $K$-shot sampling strategy that each event type has (at least) $K$ samples. Since our sampling is at sentence-level and each sentence could have multiple events, the sampling is NP-complete\footnote{The \textit{Subset Sum Problem}, a classical NP-complete problem, can be reduced to this sampling problem.} and unlikely to find a practical solution satisfying exactly $K$ samples for each event type. Therefore, we follow \citet{yang-katiyar-2020-simple} and \citet{ma-etal-2022-label} and adopt a greedy sampling algorithm to select sentences, as shown in Alg.~\ref{alg:sample algorithm}. Note that the actual sample number of each event type can be larger than $K$ under this sampling strategy. The statistics of the curated datasets are listed in Table~\ref{tab:fewshot dataset} (top).

\begin{algorithm}
\caption{Greedy Sampling}
\label{alg:sample algorithm}
    \begin{algorithmic}[1]
    \Require shot number $K$, original full dataset $\mathcal{D} = \{(\mathbf{X}, \mathbf{Y})\}$ tagged with label set $E$
    
    \State Sort $E$ based on their frequencies in $\{\mathbf{Y}\}$ as an ascending order
    \State $S \gets \phi $, $\text{Counter} \gets \text{dict}() $ 
    \For{$y \in E$}
    \State $\text{Counter}(y) \gets 0$
    \EndFor
    
    \For{$y \in E$}
    \While{$\text{Counter}(y) < K$}
    \State Sample $(\mathbf{X},\mathbf{Y}) \in \mathcal{D}$ s.t.$\exists j, y_j=y$
    \State $\mathcal{D} \gets \mathcal{D}\backslash(\mathbf{X},\mathbf{Y})$
    \State Update Counter (not only $y$ but all event types in $\mathbf{Y}$)
    \EndWhile
    \EndFor
    
    \For{$s \in \mathcal{S}$}
    \State $\mathcal{S} \gets \mathcal{S}\backslash s$ and update Counter
    \If{$\exists y \in E$, s.t. $\text{Counter}(y) < K$}
    \State $\mathcal{S} \gets \mathcal{S} \bigcup s$
    \EndIf
    \EndFor
    
    \State \Return $\mathcal{S}$
    \end{algorithmic}
\end{algorithm}

\begin{table}
\centering
\small
\caption{The statistics of curated datasets for few-shot ED tasks. Top: Low-resource setting. Bottom: Class transfer setting. We set different random seeds and generate 10 few-shot sets for each setting. We report their average statistics.
} 
\label{tab:fewshot dataset}        
\setlength{\tabcolsep}{2.5pt}
\renewcommand{\arraystretch}{0.8}
    \begin{tabular}{@{}cc|crrr} 
    \toprule
    \multicolumn{2}{c|}{\textbf{Low-resource}} & \# Labels  &\# Sent & \# Event & \# Avg shot \\ \midrule

    \multirow{3}{*}{ACE05} & 2-shot & \multirow{3}{*}{33} & 47.7 & 76.4 & 2.32 \\
    & 5-shot &  & 110.7 & 172.2 & 5.22 \\
    & 10-shot &  & 211.5 & 317.5 & 9.62\\
    \midrule
    \multirow{3}{*}{MAVEN} & 2-shot & \multirow{3}{*}{168} & 152.6 & 530.1 & 3.16 \\
    & 5-shot &  & 359.6 & 1226.3 & 7.30 \\
    & 10-shot &  & 705.1 & 2329.2 & 13.86 \\
    \midrule
    \multirow{3}{*}{ERE} & 2-shot & \multirow{3}{*}{38} & 43.6 & 108.9 & 2.87 \\
    & 5-shot & & 102.5 & 249.9 & 6.58 \\
    & 10-shot &  & 197.1 & 472.3 & 12.43 \\
    \bottomrule 
    \end{tabular}
    
    \begin{tabular}{@{}cc|crrr} 
    \toprule
    \multicolumn{2}{c|}{\textbf{Class-transfer}} & \# Labels  &\# Sent & \# Event & \# Avg shot \\ \midrule
    \multirow{3}{*}{ACE05} & 2-shot & \multirow{3}{*}{23} & 37.1 & 50.2 & 2.18\\
    & 5-shot &  & 84.6 & 113.0 & 4.91 \\
    & 10-shot &  & 159.8 & 209.9 & 9.13 \\
    \midrule
    \multirow{3}{*}{MAVEN} & 2-shot & \multirow{3}{*}{48} & 84.3 & 97.4 & 2.03 \\
    & 5-shot &  & 211.3 & 236.6 & 4.93 \\
    & 10-shot &  & 417.3 & 453.6 & 9.45 \\
    \midrule
    \multirow{3}{*}{ERE} & 2-shot & \multirow{3}{*}{28} & 39.7 & 66.1 & 2.36 \\
    & 5-shot & & 95.0 & 153.5 & 5.48 \\
    & 10-shot &  & 182.5 & 291.0 & 10.39 \\
    \bottomrule 
    \end{tabular}
\end{table}

\noindent{\textbf{Class-Transfer setting}}
This setting has a more complicated curation process, and roughly consists of two sub-steps: (1) Dividing both event types and sentences in the original dataset into two disjoint parts named source dataset and target dataset pool. (2) Using the entire source dataset, and selecting few-shot samples from the target pool to construct target set.

For step (1), we follow \citet{huang-etal-2018-zero} and  \citet{chen-etal-2021-honey} to pick out the most frequent 10, 120, and 10 event types from ACE05, MAVEN and ERE dataset respectively, as $E^{(S)}$. The remaining types are $E^{(T)}$. Then we take sentences containing any annotations in $E^{(T)}$ to $D^{(T)}_{full}$ for enriching the sampling pool of target dataset as much as possible,

\begin{equation}
\small
\nonumber
    D^{(T)}_{full} = \{(\boldsymbol{X}, R(\boldsymbol{Y}; E^{(S)})) | (\boldsymbol{X}, \boldsymbol{Y}) \in D, \exists y_j \in E^{(T)}\}
\end{equation}

where $R(\boldsymbol{Y}; E^{(S)}$ represents the relabeling operation that substituting any $y_j \in E^{(S)})$ to \texttt{N.A.} to avoid information leakage. The remaining sentences are collected as $D^{(S)}$.

\begin{equation}
\nonumber
\small
    D^{(S)} = \{(\boldsymbol{X}, R(\boldsymbol{Y}; E^{(T)})) | (\boldsymbol{X}, \boldsymbol{Y}) \notin D^{(T)}_{full} \}
\end{equation}

For step (2), we adopt the same strategy as low-resource setting to sample $K$-shot $D_{train}^{(T)}$ and $D_{dev}^{(T)}$ from target sampling pool $D^{(T)}_{full}$. Statistics of curated datasets are summarized in Table~\ref{tab:fewshot dataset} (bottom).

\subsection{Existing methods}
\label{subsec:ten methods appendix}
We conduct our empirical study on twelve representative existing methods. Besides vanilla fine-tuning and in-context learning, five of them are prompt-based and the other five are prototype-based.

\noindent{\textbf{1. Prompt-based methods}} leverage the rich knowledge in PLMs by converting specific downstream tasks to the formats that PLMs are more familiar with. We give  examples about prompt format of the five prompt-based methods in Table~\ref{tab: diff prompt}.

\noindent{\textbf{EEQA/EERC~\cite{du-cardie-2020-event, liu-etal-2020-event}:}} a QA/MRC-based method which first extracts the trigger word with a natural language query then classifies its type with an additional classifier. 

\noindent{\textbf{EDTE~\cite{lyu-etal-2021-zero}:}} a NLI-based method which enumerates all event types and judges whether a clause is entailed by any event. The clause is obtained by SRL processing and the trigger candidate is the predicate of each clause.

\noindent{\textbf{PTE~\cite{schick-schutze-2021-just}:}} a cloze-style prompt method which enumerates each word in the sentence and predicts whether it is the trigger of any event type.
% \footnote{We also notice a recently-proposed cloze-style method PILED~\cite{Li_2022_PILED},  that is specially designed for few-shot ED task. However, the authors have not  released their source code and we fail to reproduce the same result reported in their paper. Thus we do not include this method to the unified comparison at the time of paper writing.}

\noindent{\textbf{UIE~\cite{lu-etal-2022-unified}:}} a generation based method that takes in a sentence and outputs a filled \textit{universal} template, indicating the trigger words and their event types in the sentence.

\noindent{\textbf{DEGREE~\cite{hsu-etal-2022-degree}:}} also adopts a generation paradigm but it enumerates all event types by designing \textit{type-specific} template, and outputs related triggers (if have).

\begin{table*}
\centering
\small
\caption{Prompt examples for different methods based on a sentence example \texttt{X}: \textit{The current government was formed in October 2000}, in which the word \textit{formed} triggering an \textit{Start-Org} event. The underline part in UIE prompt is their designed Structured Schema Instructor (SSI), and the \textit{DESCRIPTION}($y$) in DEGREE prompt is a description about event type $y \in E$ written in natural languages. We refer readers for their original paper in details.}
    \centering
    \begin{threeparttable}
    \begin{tabular}{c|c|c}
    \toprule
    \textbf{Method}  & \textbf{Prompt Input} & \textbf{Output} \\
    \midrule
    EEQA~\scriptsize{\cite{du-cardie-2020-event}}  & \texttt{X}. What is the trigger in the event? & formed. \\
    \midrule
    \multirow{3}{*}{\shortstack{EDTE\\~\scriptsize{\cite{lyu-etal-2021-zero}}}} & Premise: \texttt{X}. Hypothesis: This text is about a Start-Org event. & Yes.\\
    & $\cdots$ & $\cdots$ \\
    & Premise: \texttt{X}. Hypothesis: This text is about an Attack event. & No. \\
    \midrule
    \multirow{3}{*}{\shortstack{PTE\\~\scriptsize{\cite{schick-schutze-2021-just}}}}  & \texttt{X}. The word \textit{formed} triggers a/an \texttt{[MASK]} event. & Start-Org \\
    & $\cdots$ & $\cdots$ \\
    & \texttt{X}. The word \textit{current} triggers a/an \texttt{[MASK]} event. & N.A. \\
    \midrule
    UIE~\scriptsize{\cite{lu-etal-2022-unified}} & \underline{<spot> Start-org <spot> Attack <spot> ... <spot>}. \texttt{X}. & (Start-Org: formed)  \\
    \midrule
    \multirow{3}{*}{\shortstack{DEGREE\\~\scriptsize{\cite{hsu-etal-2022-degree}}}}  & \texttt{X}. \textit{DESCRIPTION}(Start-Org). Event trigger is \texttt{[MASK]}. & Event trigger is formed\\
    &  $\cdots$ & $\cdots$ \\
    &  \texttt{X}. \textit{DESCRIPTION}(Attack). Event trigger is \texttt{[MASK]}. & Event trigger is N.A. \\

    \bottomrule
    \end{tabular}
    \end{threeparttable}
    \label{tab: diff prompt}
\end{table*}

\noindent{\textbf{2. Prototype-based methods}} predict an event type for each word or span by measuring the representation proximity between the samples and the \textit{prototypes} for each event type. 

\noindent{\textbf{Prototypical Network~\cite{Snell_2017}:}} a classical prototype-based method originally developed for episode learning. \citet{huang-etal-2021-shot} adapt it to low-resource setting via further splitting the training set into support set $\mathcal{S}_y$ and query set $\mathcal{Q}_y$. The prototype $\bar{c}_y$ of each event type is constructed by averaged PLM representations of samples in $\mathcal{S}_y$.
\begin{equation}
\nonumber
    h_{\bar{c}_y} = \frac{1}{\mathcal{S}_y} \sum_{s \in \mathcal{S}_y} h_s 
\end{equation}
For samples $x$ in $\mathcal{Q}_y$ during training, or in the test set during inference, $\text{logits}(y|x)$ is defined as the negative euclidean distance between $h(x)$ and $\bar{c}_y$.
\begin{equation}
\nonumber
    \text{logits}(y|x) = - ||h_x - h_{\bar{c}_y}||_2
\end{equation}

\noindent{\textbf{L-TapNet-CDT~\cite{hou-etal-2020-shot}:}} a ProtoNet-variant method with three main improvements: (1) it introduces TapNet, a variant of ProtoNet. TapNet's main difference from ProtoNet lies in a projection space $\mathcal{M}$ analytically constructed. The distance is computed in the subspace spanned by $\mathcal{M}$.
\begin{equation}
\nonumber
    \text{logits}(y|x) = - ||\mathcal{M}(h_x - h_{\bar{c}_y})||_2
\end{equation}
(2) the basis in column space of $\mathcal{M}^\perp$ is aligned with label semantic, thus $\mathcal{M}(E)$ is label-enhanced. (3) a collapsed dependency transfer (CDT) module is used solely during inference stage to scale the event-type score.
\begin{equation}
\nonumber
    \text{logits}(y|x) \gets \text{logits}(y|x) + \text{TRANS}(y)
\end{equation}

\noindent{\textbf{PA-CRF~\cite{cong-etal-2021-shot}:}} a ProtoNet-variant method with a CRF module as well. Different from CDT, however, the transition scores are approximated between event types based on the their prototypes and learned during training.

\noindent{\textbf{FSLS~\cite{ma-etal-2022-label}:}} a recently proposed few-shot NER method that generalizes well to ED task. The prototype of each event type is not constructed from support set $\mathcal{S}_y$ but from the label semantic, i.e. the PLM representation of the label name.
\begin{equation}
\begin{aligned}
\nonumber
    & e_y = \text{Event\_name}(y) \\
    & \text{logits}(y|x) = h_x^Th_{e_y}
\end{aligned}
\end{equation}

\noindent{\textbf{CONTAINER~\cite{das-etal-2022-container}:}} a contrastive learning approach. We view it as a \textit{generalized} Prototype-based method since both of their motivations are to pull together the representations of samples with same event types. Different from ProtoNet, there is no explicit division between support set and query set during training process. Instead each sample acts as query and other samples as support samples. For example, given sample $x$ with event type $e$, its \textit{special} supported set can be viewed as:
\begin{equation}
\nonumber
\mathcal{S}_y(x) = \{x'|(x', y') \in D, y'=y, x' \neq x \}
\end{equation}
Then its score related to $e$ is calculated as the average distance with samples in $\mathcal{S}_y(x)$.
\begin{equation}
\nonumber
     \text{logits}(y|x) = \sum_{x' \in \mathcal{S}_{y}(x)} \frac{-d(f(h_x), f(h_{x'}))}{|\mathcal{S}_{y}(x)|}
\end{equation}

\subsection{Implementation Details}
\label{subsec:implementation details}
For all methods, we initialize their pre-trained weights and further train them using Huggingface library.\footnote{https://huggingface.co/} Each experiment is run on single NVIDIA-V100 GPU, and the final reported performance for each setting (e.g., ACE 2-shot) is the averaged result w.r.t ten distinct few-shot training datasets which are sampled with different random seeds. We further detail the implementation of all methods.

\noindent{\textbf{1. In-context learning}}

\begin{tcolorbox}
\small
    \hspace*{-1em}{\textcolor{red}{\large{Instruction}}}
    
    \hspace*{-1em}\underline{\textbf{User}} Assume you are an event-instance annotator. Given a sentence, you need to (1) identify the word or phrase triggering the event in the sentence, and (2) classify its event type. The possible event types are listed as below: Transport, Attack, Injure, \ldots
    
    \hspace*{-1em}\underline{\textbf{ChatGPT}} Yes, I understand. 
    
    \hspace*{-1em}\underline{\textbf{User}} Please note that your annotation results must follow such format: ``Answer: ([Event\_1], [Trigger\_1]) <SEP> ([Event\_2], [Trigger\_2]) <SEP> \ldots''. If you do not find any event in this sentence, just output ``Answer: No events found.''

     \hspace*{-1em}\underline{\textbf{ChatGPT}} No problem. Let's start!

-----------------------------------------------------------

    \hspace*{-1em}{\textcolor{red}{\large{Demonstration}}}

    \hspace*{-1em}\underline{\textbf{User}} Sentence: He 's retired with the US Army.

    \hspace*{-1em}\underline{\textbf{ChatGPT}} Answer: (End-Position, retired)

    \hspace*{-1em}\underline{\textbf{User}} Sentence: I wouldn't be surprised if this Arroyo character (who was killed by police as he tried to flee) had a serious criminal past but has been allowed to walk free due to our lax judicial process.

    \hspace*{-1em}\underline{\textbf{ChatGPT}} Answer: (Die, killed) <SEP> (Acquit, walk free)

    \hspace*{-1em}\underline{\textbf{User}} Sentence: Granting a local sheik power to set up an administrative committee representing the groups in the region.

    \hspace*{-1em}\underline{\textbf{ChatGPT}} Answer: No events found.

-----------------------------------------------------------

    \hspace*{-1em}{\textcolor{red}{\large{Question}}}

    \hspace*{-1em}\underline{\textbf{User}} Hariri submitted his resignation during a 10-minute meeting with the head of state at the Baabda presidential palace, outside the capital.
    
\end{tcolorbox}

We use ChatGPT (\texttt{gpt-3.5-turbo-0301}) provided by OpenAI APIs~\footnote{https://platform.openai.com/docs/api-reference} for in-context learning. The prompt simulates and records the chatting history between the \underline{\textbf{user}} and the \underline{\textbf{model}}. We show one example as above. The prompt consists of three parts: (1) the instruction telling LLMs the task purposes and input-output formats, (2) the demonstration showcasing several input-output pairs to teach LLMs the task and (3) the input of test instance. We feed the prompt into LLMs and expect them to generate extracted answers. Specifically, we set the temperature as 0 and maximum output token as 128. We make all samples in few-shot train set as demonstration samples if their total length is smaller than the maximum input token length (4096). Otherwise we retrieve similar demonstration samples for each test instance to fill up the input prompt.
The similarity between two instances are measured from their embeddings~\cite{gao-etal-2021-simcse}. For MAVEN dataset, we further sample a test subset, with 1000 instances, from the original one for our evaluation.

\noindent{\textbf{2. Prompt-based methods}}
We keep all other hyperparameters the same as in their original papers, except learning rates and epochs. We grid-search best learning rates in [1e-5, 2e-5, 5e-5, 1e-4] for each setting. As for epochs, we find the range of appropriate epochsis highly affected by the prompt format. Therefore we search for epochs method by method without a unified range.

\noindent{\textbf{EEQA~\cite{du-cardie-2020-event}:}} We use their original code\footnote{https://github.com/xinyadu/eeqa} and train it on our datasets.

\noindent{\textbf{EDTE~\cite{lyu-etal-2021-zero}:}}  We use their original code\footnote{https://github.com/veronica320/Zeroshot-Event-Extraction} and train it on our datasets.

\noindent{\textbf{PTE~\cite{schick-schutze-2021-just}:}} We implement this method on OpenPrompt~\cite{ding-etal-2022-openprompt}.

\noindent{\textbf{UIE~\cite{lu-etal-2022-unified}:}}  We use their original code\footnote{https://github.com/universal-ie/UIE} and train it on our datasets.

\noindent{\textbf{DEGREE~\cite{hsu-etal-2022-degree}:}} We reproduce this method based on their original code\footnote{https://github.com/PlusLabNLP/DEGREE} and train it on our datasets. And we drop event keywords not occurring in few-shot training dataset from prompt to avoid information leakage.

\noindent{\textbf{3. Prototype-base methods}} We build a codebase based on the unified view. We then implement these methods directly on the unified framework, by having different choices for each design element. To ensure the correctness of our codebase, we also compare between results obtained from our implementation and original code for each method, and find they achieving similar performance on few-shot ED datasets. 

For all methods (including  \textit{unified baseline}), we train them with the AdamW~\cite{Loshchilov2017DecoupledWD} optimizer with linear scheduler and 0.1 warmup step. We set weight-decay coefficient as 1e-5 and maximum gradient norms as 1.0. We add a 128-long window centering on the trigger words and only encode the words within the window; in other words, the maximum encoding sequence length is 128. The batch size is set as 128, and training steps as 200 if the transfer function is scaled (see Section~\ref{subsec: analyze on prototype-based methods, low-resource settings}) otherwise 500. We grid-search best learning rates in [1e-5, 2e-5, 5e-5, 1e-4] for each setting. For ProtoNet and its variants, we further split the sentences into support set and query set. The number in support set $K_S$ and query set $K_Q$ are (1, 1) for 2-shot settings, (2, 3) for 5-shot settings. The split strategy is (2, 8) for 10-shot dataset constructed from MAVEN and (5, 5) for others. For methods adopting MoCo-CL setting (also see Section~\ref{subsec: analyze on prototype-based methods, low-resource settings}), we maintain a queue storing sample representations with length 2048 for ACE/ERE 2-shot settings and 8192 for others. For methods adopting CRF, we follow default hyperparameters about CRF in their original papers. For methods adopting scaled transfer functions, we grid search the scaled coefficient $\tau$ in [0.1, 0.2, 0.3].
\section{Low-resource Setting-Extended}
\label{sec:supplementary for low-resource setting}

\subsection{Transfer function and Distance function}
\label{subsec:appendix d and f}

We consider several combinations about distance and transfer functions listed in Table~\ref{tab:d-f-combination}. We choose cosine similarity (S), negative euclidean distance (EU) and their scaled version (SS/SEU) as distance functions. And we pick out identify (I), down-projection (D) and their normalization version (N/DN) as transfer function. We additionally consider the KL-reparameterization combination (KL-R) used in CONTAINER.

\begin{table}[htbp!]
\caption{Variants on distance function $d(u, v)$ (top) and transfer function $f(h)$ (bottom).}
\label{tab:d-f-combination}
    \centering
        \small
        \begin{tabular}{@{}l|c@{}} 
            \toprule
            \textbf{Distance function} & $d(u, v)$  \\
            \midrule
            Cosine similarity (S) & $u^Tv$ \\
            Scaled cosine similarity (SS) & $u^Tv/\tau$\\
            JS Divergence (KL) & $\text{JSD}(u||v)$ \\
            Euclidean distance (EU) & $-||u - v||_2$ \\
            Scaled euclidean distance (SEU) & $-||u - v||_2/\tau$ \\
            \midrule\midrule 
            \textbf{Transfer function} & $f(h)$  \\
            \midrule
            Identify (I) & $h$ \\
            Down-projection (D) & $\mathcal{M}h$ \\
            Reparameterization (R) & $\mathcal{N}(\mu(h), \Sigma(h))$ \\
            Normalization (N) & $h/||h||$ \\
            Down-projection + Normalization (DN) & $\mathcal{M}h/||h||$ \\
            \bottomrule 
            \end{tabular}
    \end{table}

\begin{figure}[htbp!]
    \includegraphics[width=0.95\linewidth]{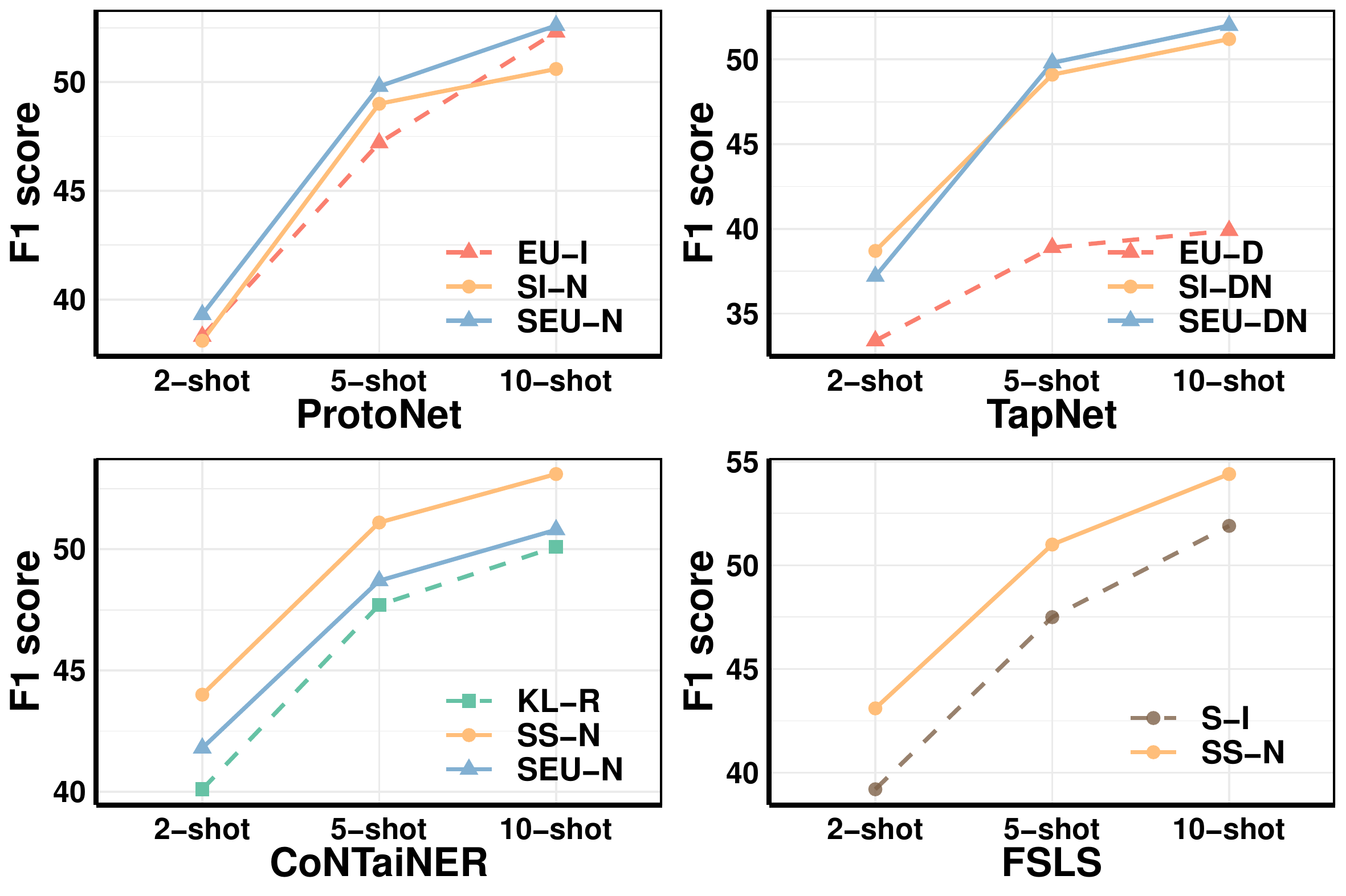}
    \caption{Performance of different $(d, f)$ combinations on ACE05.}
    \label{fig:d and f}
\end{figure}

We conduct experiments with four existing prototype-based methods\footnote{We \textit{degrade} L-TapNet-CDT to TapNet, and do not include PA-CRF here, because CRF and label-enhancement are not the factors considered in this subsection.} by only changing their transfer and distance functions. We illustrate their results on ACE dataset in Figure~\ref{fig:d and f}. (1) From comparison about performance in ProtoNet and TapNet, we find TapNet, i.e., the down-projection transfer, shows no significant improvement on few-shot ED tasks. (2) A scaled coefficient in distance function achieves strong performance with normalization transfer function, while the performance collapses (failing to converge) without normalization. (3) For ProtoNet and TapNet, scaled euclidean distance (SEU) is a better choice for distance function, while other methods prefer scaled cosine similarity (SS). Based on the findings above, we substitute $d$ and $f$ to the most appropriate for all existing methods and observe a significant improvement on all three datasets, as shown in Table~\ref{tab:d and f}.

\begin{table*}[htbp!]
 \setlength\tabcolsep{3pt}
 \centering
 \small
 \caption{Performance comparison of methods  w/ and w/o adjustment on distance function $d$ and transfer function $f$. The most appropriate distance functions are scaled euclidean distance (SEU) for ProtoNet and TapNet and scaled cosine similarity (SS) for other two. The most appropriate transfer function is normalization (N) for all four existing methods. The results are averaged among 10 repeated experiments and sample standard deviations are in round brackets. We highlight the better one for each method \textit{w/} and \textit{w/o} adjustment.}
    \begin{threeparttable}
        \begin{tabular}{ll|ccc|ccc|ccc} 
        \toprule
        \multicolumn{2}{c|}{\textbf{Methods}} & \multicolumn{3}{c|}{\textbf{ACE05}}  & \multicolumn{3}{c|}{\textbf{MAVEN}} & \multicolumn{3}{c}{\textbf{ERE}} \\
        & & {2-shot} & {5-shot} & {10-shot} & {2-shot} & {5-shot} & {10-shot}  & {2-shot} & {5-shot} & {10-shot} \\
        \midrule
        \multirow{2}{*}{ProtoNet} & w/o adjust & $38.3${\tiny $(5.0)$}  & $47.2${\tiny $(3.9)$} & $52.3${\tiny $(2.4)$} & $44.5${\tiny $(2.2)$}  & $51.7${\tiny $(0.6)$} & $55.4${\tiny $(0.2)$} & $31.6${\tiny $(2.7)$} & $39.7${\tiny $(2.4)$} & $\textbf{44.3}${\tiny $(2.3)$} \\
         & w/ adjust & $\textbf{39.3}${\tiny $(4.6)$} & $\textbf{49.8}${\tiny $(4.3)$} & $\textbf{52.6}${\tiny $(1.9)$} & $\textbf{46.7}${\tiny $(1.6)$} & $\textbf{52.8}${\tiny $(0.6)$} & $\textbf{56.5}${\tiny $(0.6)$} & $\textbf{32.6}${\tiny $(3.0)$} & $\textbf{40.1}${\tiny $(1.9)$} &  ${44.2}${\tiny $(1.9)$} \\
         \midrule
        \multirow{2}{*}{TapNet} & w/o adjust & $\textbf{38.7}${\tiny $(4.3)$} & ${49.1}${\tiny $(4.5)$} & ${51.2}${\tiny $(1.7)$} & ${45.7}${\tiny $(1.8)$} & ${51.7}${\tiny $(1.1)$} & ${55.0}${\tiny $(0.7)$} & ${35.3}${\tiny $(3.8)$} & ${40.2}${\tiny $(2.5)$} & ${44.7}${\tiny $(2.9)$} \\
         & w/ adjust & ${37.2}${\tiny $(5.6)$} & $\textbf{49.8}${\tiny $(3.1)$} & $\textbf{52.0}${\tiny $(1.9)$} & $\textbf{46.1}${\tiny $(1.9)$} & $\textbf{51.9}${\tiny $(0.6)$} & ${55.0}${\tiny $(0.6)$} & $\textbf{37.0}${\tiny $(4.0)$} & $\textbf{43.4}${\tiny $(1.9)$} & $\textbf{46.4}${\tiny $(2.9)$} \\
        \midrule
        \multirow{2}{*}{CONTAINER} & w/o adjust & ${40.1}${\tiny $(3.8)$} & ${47.7}${\tiny $(3.3)$} & ${50.1}${\tiny $(1.8)$} &  $44.2${\tiny $(1.4)$} &  $50.8${\tiny $(0.9)$} &  $52.9${\tiny $(0.3)$} & $34.4${\tiny $(3.6)$} & $39.3${\tiny $(1.9)$} & $44.5${\tiny $(2.3)$} \\
         & w/ adjust & $\textbf{44.0}${\tiny $(3.2)$} & $\textbf{51.1}${\tiny $(1.1)$} & $\textbf{53.1}${\tiny $(1.8)$} & $\textbf{44.6}${\tiny $(1.7)$} & $\textbf{52.1}${\tiny $(0.5)$} & $\textbf{55.1}${\tiny $(0.4)$} & $\textbf{36.5}${\tiny $(4.1)$} & $\textbf{42.0}${\tiny $(1.9)$} & $\textbf{45.4}${\tiny $(1.5)$} \\
         \midrule
        \multirow{2}{*}{FSLS} & w/o adjust & $39.2${\tiny $(3.4)$} & $47.5${\tiny $(3.2)$}  & $51.9${\tiny $(1.7)$} & $46.7${\tiny $(1.2)$} & $51.5${\tiny $(0.5)$} & $\textbf{56.2}${\tiny $(0.2)$} & $34.5${\tiny $(3.1)$} & $39.8${\tiny $(2.5)$} & $44.0${\tiny $(2.0)$}\\
         & w/ adjust & $\textbf{43.1}${\tiny $(3.4)$} & $\textbf{51.0}${\tiny $(2.4)$} & $\textbf{54.4}${\tiny $(1.5)$} & $\textbf{48.3}${\tiny $(1.6)$} & $\textbf{53.4}${\tiny $(1.6)$} & ${56.1}${\tiny $(0.7)$} & $\textbf{35.7}${\tiny $(2.1)$} & $\textbf{40.6}${\tiny $(2.4)$} & $\textbf{45.4}${\tiny $(1.7)$} \\
        \bottomrule
        \end{tabular}
    \end{threeparttable}
    \label{tab:d and f}
\end{table*}

\subsection{CRF module}
\label{subsec:appendix CRF module}
\begin{figure}[htbp!]
\centering
    \subfigure[ACE05]{
    \includegraphics[width=\linewidth]{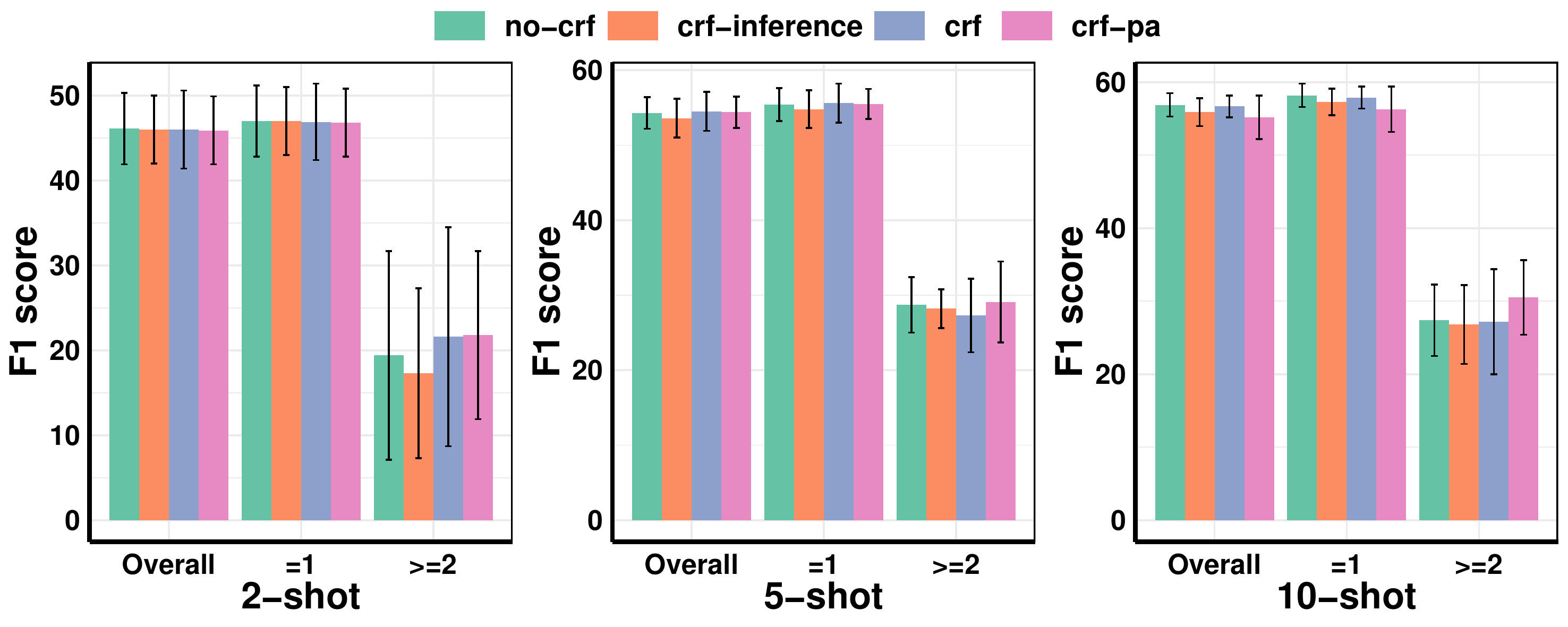}
    }%
    
    \subfigure[MAVEN]{
    \centering
    \includegraphics[width=\linewidth]{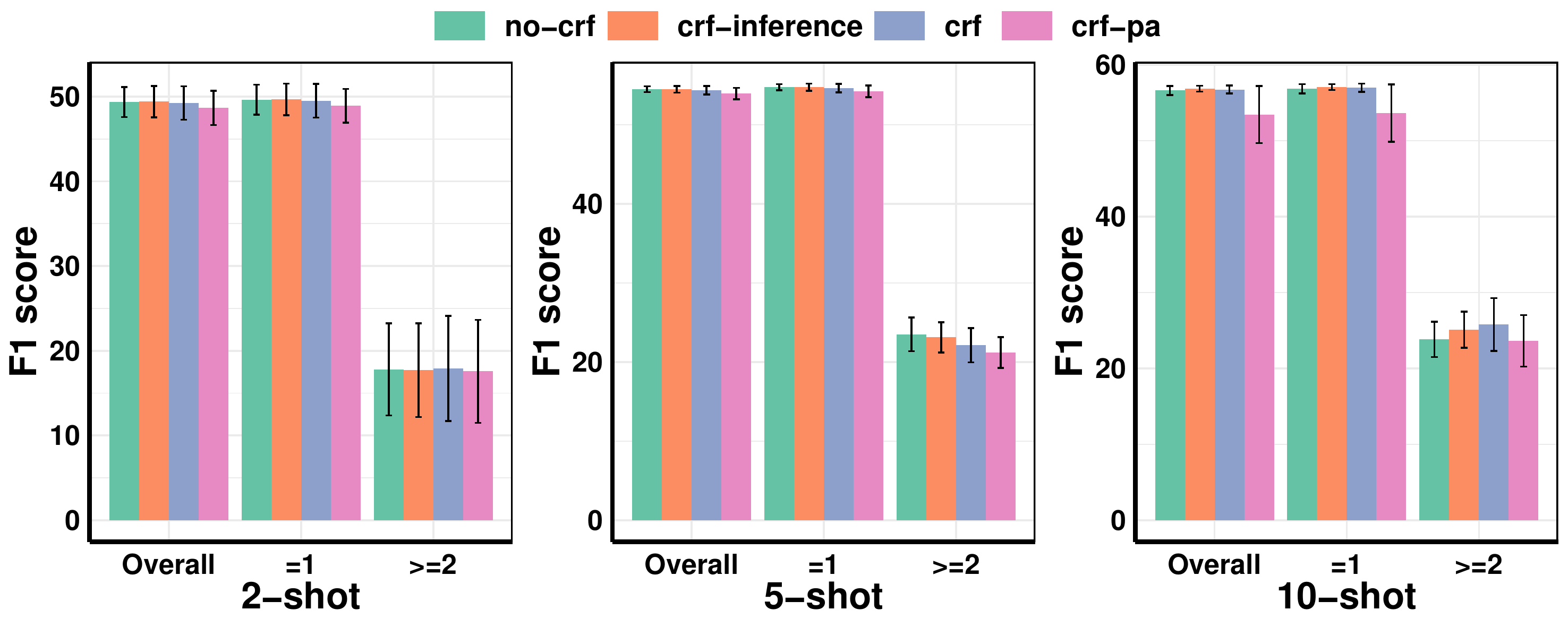}
    }%
\caption{Overall performance of different CRF variants on ACE05 and MAVEN datasets. We also provide performance grouped by trigger word length: \textbf{$=1$}: single trigger words. \textbf{$\geq 2$}: trigger phrases.}
\label{fig:crf}
\end{figure}

We explore whether CRF improves the performance of few-shot ED task. Based on \textit{Ll-MoCo} model we developed in Section~\ref{subsec: analyze on prototype-based methods, low-resource settings}, we conduct experiment with three different CRF variants, CDT (CRF inference~\citealt{hou-etal-2020-shot}), vanilla CRF~\cite{2001_CRF} and PA-CRF~\cite{cong-etal-2021-shot}, on ACE05 and MAVEN datasets. Their results are in Figure~\ref{fig:crf}. It shows different CRF variants achieve similar result compared with model without CRF, while a trained CRF (and its prototype-enhanced variant) slightly benefits multiple-word triggers  when the sample is extremely scarce (see ACE05 2-shot). These results are inconsistent with other similar sequence labeling tasks such as NER or slot tagging, in which CRF usually significantly improves model performance. We speculate it is due to that the pattern of triggers in ED task is relatively simple. To validate such assumption, we count all triggers in ACE05 and MAVEN datasets. We find that above $96\%$ of triggers are single words, and most of the remaining triggers are verb phrases 
% (only about $0.5\%$ of triggers are phrases having three or more words with complicated structure). 
Thus the explicit modeling of transfer dependency among different event types is somewhat not very meaningful under few-shot ED task. Hence, we drop CRF module in the \textit{unified baseline}.

\subsection{Prototype source}
We discuss the benefit of combining two kinds of prototype sources in Section~\ref{subsec: analyze on prototype-based methods, low-resource settings}, i.e., label semantic and event mentions, and show some results in Figure~\ref{fig:aggregation form}. Here we list full results on all three datasets in Table~\ref{tab:aggregation form}. The results further validate our claims: (1) leveraging both label semantics and mentions as prototype sources improve  performance under almost all settings. (2) Merging the two kinds of sources at the loss-level is the best choice among the three aggregation alternatives.

\begin{table*}
 \centering
 \small
 \caption{Performance with different (1) prototype sources and (2) aggregation form. \textbf{ProtoNet}: only event mentions. \textbf{FSLS}: label semantic. \textbf{Lf-ProtoNet}: aggregate two types of prototype sources at feature-level. \textbf{Ls-ProtoNet}: at score-level. \textbf{Ll-ProtoNet}: at loss-level. The results are averaged over 10 repeated experiments and sample standard deviations are in round brackets.}
    \begin{threeparttable}
        \begin{tabular}{l|ccc|ccc|ccc} 
        \toprule
        Methods & \multicolumn{3}{c|}{\textbf{ACE05}}  & \multicolumn{3}{c|}{\textbf{MAVEN}} & \multicolumn{3}{c}{\textbf{ERE}} \\
        & {2-shot} & {5-shot} & {10-shot} & {2-shot} & {5-shot} & {10-shot}  & {2-shot} & {5-shot} & {10-shot} \\
        \midrule
        ProtoNet & ${39.3}${\tiny $(4.6)$} & ${49.8}${\tiny $(4.3)$} & ${52.6}${\tiny $(1.9)$} & ${46.7}${\tiny $(1.6)$} & ${52.8}${\tiny $(0.6)$} & ${56.0}${\tiny $(0.6)$} & ${32.6}${\tiny $(3.0)$} & ${40.1}${\tiny $(1.9)$} &  ${44.2}${\tiny $(1.9)$}  \\
        FSLS &  $\underline{43.0}${\tiny $(3.4)$} & ${50.6}${\tiny $(2.4)$} & $\textbf{54.1}${\tiny $(1.5)$} & ${48.3}${\tiny $(1.6)$} & ${53.4}${\tiny $(0.2)$} & ${56.1}${\tiny $(0.7)$} & ${35.7}${\tiny $(2.1)$} & ${40.6}${\tiny $(2.4)$} & $\textbf{45.4}${\tiny $(1.7)$} \\
        Lf-ProtoNet & $41.9${\tiny $(3.8)$} &  $50.8${\tiny $(3.0)$} &  $52.9${\tiny $(2.4)$} & ${49.0}${\tiny $(1.1)$} & ${53.4}${\tiny $(1.0)$} & ${56.3}${\tiny $(0.7)$} & ${35.3}${\tiny $(3.6)$} & $\underline{41.8}${\tiny $(1.8)$} & ${45.3}${\tiny $(2.2)$}  \\
        Ls-ProtoNet & ${42.7}${\tiny $(4.8)$} & $\textbf{51.2}${\tiny $(2.9)$} & ${52.7}${\tiny $(1.7)$} & $\underline{49.3}${\tiny $(1.9)$} & $\underline{53.5}${\tiny $(0.7)$} & $\underline{56.5}${\tiny $(0.1)$} & $\underline{36.0}${\tiny $(2.5)$} & ${41.3}${\tiny $(3.6)$} & ${44.8}${\tiny $(2.5)$}  \\
        Ll-ProtoNet & $\textbf{43.3}${\tiny $(4.0)$} & $\underline{50.9}${\tiny $(2.7)$} & $\underline{53.0}${\tiny $(2.1)$} & $\textbf{50.2}${\tiny $(1.5)$} & $\textbf{54.3}${\tiny $(0.8)$} & $\textbf{56.7}${\tiny $(0.6)$} & $\textbf{37.6}${\tiny $(3.1)$} & $\textbf{43.0}${\tiny $(2.4)$} & $\underline{45.3}${\tiny $(1.9)$}    \\
        \bottomrule
        \end{tabular}
    \end{threeparttable}
    \label{tab:aggregation form}
\end{table*}

\subsection{Contrastive Learning}
\label{subsec:appendix CL}
Contrastive Learning (CL~\citealt{2006_CL}) is initially developed for self-supervised representation learning and is recently used to facilitate supervised learning as well. It pulls samples with same labels together while pushes samples with distinct labels apart in their embedding space. We view CL as a \textit{generalized} format of prototype-based methods and include it to the unified view. Under such view, every sample is a prototype and each single event type could have multiple prototypes. Given an event mention, its distances to the prototypes are computed and aggregated by event types to determine the overall distance to each event type. 

\noindent{\textbf{Two types of Contrastive Learning}}

We name the \textbf{representation} of event mention as query and prototypes (i.e., other event mentions) as keys. Then CL could be further split into two cases, in-batch CL~\cite{2020_simclr} and MoCo CL~\cite{2020_MoCo}, according to where their \textbf{keys} are from. In-batch CL views other event mentions within the same batch as the keys, and the encoder for computing the queries and keys in batch-CL is updated end-to-end by back-propagation. For MoCo CL, the encoder for key is momentum-updated along the encoder for query, and it accordingly maintains a queue to store keys and utilizes them multiple times once they are previously computed. We refer readers to MoCo CL~\cite{2020_MoCo} for the details of in-batch CL and MoCo CL.

CONTAINER~\cite{das-etal-2022-container} adopts in-batch CL setting for few-shot NER model and we transfer it to ED domain in our empirical study. We further compare the two types of CL for our \textit{unified baseline} with effective components in Section~\ref{subsec: analyze on prototype-based methods, low-resource settings} and present the full results in Table~\ref{tab:in-batch or moco}. We observe in-batch CL outperforms MoCo-CL when the number of the sentence is small, and the situation reverses with the increasing of sentence number. We speculate it is due to two main reasons: (1) When all sentences could be within the single batch, in-batch CL is a better approach since it computes and updates all representations of keys and queries end-to-end by back propagation, while MoCo-CL computes the key representation by a momentum-updated encoder with gradient stopping. When the sentence number is larger than batch size, however, in-batch CL lose the information of some samples in each step, while MoCo-CL keeps all samples within the queue and leverages these approximate representations for a more extensive comparison and learning. (2) MoCo-CL also has an effect of data-augmentation under few-shot ED task, since the sentence number is usually much smaller than the queue size. Then the queue would store multiple representations for each sample, which are computed and stored in different previous steps. The benefits of such data augmentation take effect when there are relatively abundant sentences and accordingly diverse augmentations.

\begin{table*}
 \centering
 \small
 \caption{Performance with three label-enhanced approaches. The number in square bracket represents (average) sentence number under this setting. Averaged F1-scores with sample standard deviations on 10 repeated experiments are shown.}
    \begin{threeparttable}
        % \begin{tabular}{l|ccc|ccc|ccc} 
        \begin{tabular}{p{2.4cm}|p{1cm}p{1cm}p{1cm}|p{1cm}p{1cm}p{1cm}|p{1cm}p{1cm}p{1cm}}
        \toprule
        \multirow{3}{*}{Method} & \multicolumn{3}{c|}{\textbf{ACE05}}  & \multicolumn{3}{c|}{\textbf{MAVEN}} & \multicolumn{3}{c}{\textbf{ERE}} \\
        & {2-shot} & {5-shot} & {10-shot} & {2-shot} & {5-shot} & {10-shot}  & {2-shot} & {5-shot} & {10-shot} \\
        & [48] & [111] & [212] & [153] & [360] & [705] & [44] & [103] & [197] \\
        \midrule
        Ll-ProtoNet & $\underline{43.3}${\tiny $(4.0)$} & ${50.9}${\tiny $(2.7)$} & ${53.0}${\tiny $(2.1)$} & $\textbf{50.2}${\tiny $(1.5)$} & ${54.3}${\tiny $(0.8)$} & ${56.7}${\tiny $(0.6)$} & ${37.6}${\tiny $(3.1)$} & ${43.0}${\tiny $(2.4)$} & ${45.3}${\tiny $(1.9)$}    \\
        Ll-CONTAINER & $\textbf{45.9}${\tiny $(3.7)$} & $\textbf{54.0}${\tiny $(2.6)$} & $\underline{55.8}${\tiny $(1.3)$} & ${49.2}${\tiny $(1.6)$} & $\underline{54.3}${\tiny $(0.6)$} & $\underline{57.3}${\tiny $(0.7)$} & $\textbf{39.5}${\tiny $(2.4)$} & $\underline{45.5}${\tiny $(2.8)$} & $\underline{46.9}${\tiny $(1.8)$}  \\
        Ll-MoCo & ${42.8}${\tiny $(4.1)$} & $\underline{53.6}${\tiny $(4.1)$} & $\textbf{56.9}${\tiny $(1.6)$} &  $\underline{49.5}${\tiny $(1.7)$}  & $\textbf{54.7}${\tiny $(0.8)$} & $\textbf{57.8}${\tiny $(1.2)$} & $\underline{38.8}${\tiny $(2.4)$} & $\textbf{46.0}${\tiny $(3.0)$} & $\textbf{48.4}${\tiny $(2.6)$}  \\
        \bottomrule
        \end{tabular}
    \end{threeparttable}
    \label{tab:in-batch or moco}
\end{table*}
\section{Class-transfer Setting-Extended}
\label{sec:supplementary for class-transfer setting}

\subsection{Prompt-based methods}
\label{subsec: appendix class transfer prompt}
We list the results of existing prompt-based methods on class-transfer setting in Table~\ref{tab:class-transfer-prompt}. See detailed analysis in Section~\ref{subsec:class-transfer-prompt}.

\subsection{Prototype-based methods}
\label{subsec: appendix class transfer prototype}
We list the results of existing prototype-based methods plus our developed \textit{unified baseline} under class-transfer setting in Table~\ref{tab:class-transfer-prototype}. Note that we substitute the appropriate distance functions $d$ and transfer functions $f$ obtained in Section~\ref{subsec: analyze on prototype-based methods, low-resource settings} for existing methods. See detailed analysis in Section~\ref{subsec:class-transfer-prototype}.

\begin{table*}
 \centering
 \setlength\tabcolsep{3pt}
 \small
 \caption{
    Prompt-based methods under class-transfer setting. Averaged F1-scores with sample standard deviations on 10 repeated experiments are shown. We also list results of \textit{w/o} and \textit{w/} transfer for comparison. 
 }
    \begin{threeparttable}
        \begin{tabular}{cr|ccc|ccc|ccc} 
        \toprule
        \multicolumn{2}{c|}{\textbf{Method}} & \multicolumn{3}{c|}{\textbf{ACE05}} & \multicolumn{3}{c|}{\textbf{MAVEN}} & \multicolumn{3}{c}{\textbf{ERE}}\\
         & & {2-shot} & 5-shot & 10-shot & 2-shot & 5-shot & 10-shot & 2-shot & 5-shot & 10-shot \\
        \midrule
        \multirow{2}{*}{EEQA} & \textit{w/o transfer} & ${17.6}${\tiny $(4.9)$} & ${33.2}${\tiny $(3.8)$} & ${41.9}${\tiny $(2.9)$} & ${14.9}${\tiny $(4.4)$} & ${44.8}${\tiny $(3.1)$} & ${53.9}${\tiny $(0.7)$} & ${19.6}${\tiny $(7.5)$} & ${36.8}${\tiny $(3.1)$} & ${44.2}${\tiny $(4.3)$}  \\
        & \textit{w/ transfer} & ${35.1}${\tiny $(8.5)$} & ${52.5}${\tiny $(6.1)$} & $\textbf{59.1}${\tiny $(2.5)$} & ${35.0}${\tiny $(4.7)$} & ${54.7}${\tiny $(1.7)$} & ${60.0}${\tiny $(0.7)$} & ${26.8}${\tiny $(5.2)$} & $39.1${\tiny $(3.1)$} & $45.9${\tiny $(2.8)$} \\
        \midrule
        \multirow{2}{*}{PTE} & \textit{w/o transfer} & ${39.7}${\tiny $(4.1)$} & ${51.1}${\tiny $(5.4)$} & ${54.5}${\tiny $(3.0)$} & ${52.0}${\tiny $(1.3)$} & $\textbf{61.0}${\tiny $(1.4)$} & ${62.5}${\tiny $(2.3)$} & $\underline{47.1}${\tiny $(4.9)$} & $\underline{51.0}${\tiny $(5.7)$} & $\underline{54.1}${\tiny $(4.1)$} \\
        & \textit{w/ transfer} & $\underline{49.1}${\tiny $(4.9)$} & $\underline{55.4}${\tiny $(5.8)$} & ${54.2}${\tiny $(4.4)$} & ${52.0}${\tiny $(2.9)$} & $\underline{60.8}${\tiny $(1.0)$} & ${61.5}${\tiny $(1.5)$} & ${42.6}${\tiny $(3.7)$} & $\textbf{51.0}${\tiny $(3.1)$} & $\textbf{55.3}${\tiny $(2.3)$} \\
        \midrule
        \multirow{2}{*}{UIE} & \textit{w/o transfer} & ${24.5}${\tiny $(3.9)$} & ${39.3}${\tiny $(3.2)$} &${40.6}${\tiny $(3.9)$} & ${25.3}${\tiny $(8.1)$} &${49.2}${\tiny $(2.2)$} &${57.4}${\tiny $(2.3)$} & ${22.9}${\tiny $(9.0)$} & ${35.1}${\tiny $(4.2)$} & ${39.3}${\tiny $(2.3)$} \\
        & \textit{w/ transfer} &${47.0}${\tiny $(5.4)$} &${54.0}${\tiny $(4.2)$} &${54.7}${\tiny $(7.3)$} & ${40.3}${\tiny $(1.7)$} & ${49.8}${\tiny $(1.6)$} & ${54.1}${\tiny $(1.5)$} & ${36.9}${\tiny $(4.6)$} & ${41.1}${\tiny $(4.2)$} & ${41.9}${\tiny $(4.6)$} \\
        \midrule
        \multirow{2}{*}{DEGREE} & \textit{w/o transfer} & ${33.4}${\tiny $(6.6)$} & ${44.2}${\tiny $(2.2)$} & ${50.5}${\tiny $(6.3)$} & $\underline{53.6}${\tiny $(1.9)$} & ${56.9}${\tiny $(5.7)$} & $\underline{63.8}${\tiny $(1.2)$} & ${39.1}${\tiny $(5.9)$} & ${41.8}${\tiny $(3.2)$} & ${43.9}${\tiny $(6.2)$} \\
        & \textit{w/ transfer} & $\textbf{52.4}${\tiny $(3.7)$}  & $\textbf{56.7}${\tiny $(4.6)$} & $\underline{59.0}${\tiny $(4.7)$} & $\textbf{54.5}${\tiny $(5.1)$}  & ${59.6}${\tiny $(6.3)$} & $\textbf{65.1}${\tiny $(2.7)$} & $\textbf{50.1}${\tiny $(3.6)$} & ${50.3}${\tiny $(2.8)$} & ${48.5}${\tiny $(2.5)$} \\
        \bottomrule
        \end{tabular}
    \end{threeparttable}
    \label{tab:class-transfer-prompt}
\end{table*}

\begin{table*}
 \centering
 \setlength\tabcolsep{3pt}
  \caption{
    Full results about prototype-based methods under class transfer setting. Averaged F1-scores with sample standard deviations on 10 repeated experiments are shown. We enumerate all possible combinations on models of source and target datasets. 
 }
 \small
    \begin{threeparttable}
        \begin{tabular}{cc|ccc|ccc|ccc} 
        \toprule
        \multicolumn{2}{c|}{\textbf{Method}} & \multicolumn{3}{c|}{\textbf{ACE05}} & \multicolumn{3}{c|}{\textbf{MAVEN}} & \multicolumn{3}{c}{\textbf{ERE}} \\
        Source & Target & 2-shot & 5-shot & 10-shot & 2-shot & 5-shot & 10-shot & 2-shot & 5-shot & 10-shot \\
        \midrule
        $\_$ & \parbox[t]{1mm}{\multirow{6}{*}{\rotatebox[origin=c]{90}{Fine-tuning}}} & ${28.1}${\tiny $(9.9)$} & ${37.0}${\tiny $(8.3)$} & ${45.8}${\tiny $(4.0)$} & ${21.2}${\tiny $(11.5)$} & ${46.6}${\tiny $(4.2)$} & ${55.3}${\tiny $(4.8)$} & ${40.4}${\tiny $(3.8)$} & ${45.9}${\tiny $(3.8)$} & ${48.2}${\tiny $(2.2)$} \\
        Fine-tuning & & ${39.1}${\tiny $(6.7)$} & ${49.5}${\tiny $(11.9)$} & ${51.4}${\tiny $(9.3)$} & ${44.4}${\tiny $(1.8)$} & ${58.3}${\tiny $(1.9)$} & ${63.0}${\tiny $(1.9)$} & ${34.1}${\tiny $(6.9)$} & ${47.0}${\tiny $(4.5)$} & ${50.0}${\tiny $(2.3)$}  \\
        CONTAINER & & ${28.7}${\tiny $(5.8)$} & ${37.4}${\tiny $(11.6)$} & ${42.7}${\tiny $(8.0)$} & ${49.4}${\tiny $(2.8)$} & ${59.3}${\tiny $(1.4)$} & ${63.6}${\tiny $(1.7)$} & ${36.3}${\tiny $(8.9)$} & ${47.3}${\tiny $(3.7)$} & ${47.3}${\tiny $(4.0)$} \\
        L-TapNet & & ${31.7}${\tiny $(5.7)$} & ${41.5}${\tiny $(4.2)$} & ${43.1}${\tiny $(2.6)$} & ${40.0}${\tiny $(1.8)$} & ${54.3}${\tiny $(1.4)$} & ${59.9}${\tiny $(1.4)$} & ${36.8}${\tiny $(4.7)$} & ${44.0}${\tiny $(5.3)$} & ${48.7}${\tiny $(2.1)$}  \\
        FSLS & & ${42.3}${\tiny $(8.5)$} & ${51.6}${\tiny $(6.9)$} & ${56.7}${\tiny $(8.6)$} & ${47.1}${\tiny $(2.7)$} & ${58.1}${\tiny $(1.1)$} & ${62.9}${\tiny $(1.6)$} & ${41.2}${\tiny $(4.7)$} & ${49.8}${\tiny $(3.6)$} & ${53.2}${\tiny $(3.4)$}    \\
        Unified Baseline & & ${39.8}${\tiny $(6.0)$} & ${47.4}${\tiny $(6.2)$} & ${54.3}${\tiny $(6.4)$} & ${48.8}${\tiny $(1.7)$} & ${58.8}${\tiny $(1.0)$} & ${63.9}${\tiny $(1.0)$} & ${39.8}${\tiny $(5.2)$} & ${46.1}${\tiny $(3.5)$} & ${50.8}${\tiny $(3.4)$}   \\
        \midrule
        $\_$ & \parbox[t]{1mm}{\multirow{6}{*}{\rotatebox[origin=c]{90}{CONTAINER}}} & ${40.1}${\tiny $(3.0)$} & ${47.3}${\tiny $(5.8)$} &  ${49.1}${\tiny $(4.7)$} & ${47.9}${\tiny $(3.5)$} &  ${63.5}${\tiny $(1.1)$} &  ${68.5}${\tiny $(2.1)$} & ${46.5}${\tiny $(4.9)$} & ${49.2}${\tiny $(3.0)$} & ${53.5}${\tiny $(3.3)$} \\
        Fine-tuning & & ${37.2}${\tiny $(9.5)$}  & ${45.0}${\tiny $(8.1)$} & ${52.7}${\tiny $(8.7)$}  &${54.3}${\tiny $(3.4)$}  & ${\underline{64.3}}${\tiny $(1.1)$} &  ${66.8}${\tiny $(2.9)$} & ${35.0}${\tiny $(4.0)$} & ${42.1}${\tiny $(4.6)$} & $47.6${\tiny $(4.0)$}   \\
        CONTAINER &  &  ${30.6}${\tiny $(5.4)$} & ${38.3}${\tiny $(5.4)$} & ${37.6}${\tiny $(4.5)$} & ${47.5}${\tiny $(6.4)$} & ${57.1}${\tiny $(3.4)$} & ${54.7}${\tiny $(2.2)$} & ${42.1}${\tiny $(4.8)$} & ${46.6}${\tiny $(4.9)$} & ${51.7}${\tiny $(2.9)$} \\
        L-TapNet &  & ${33.0}${\tiny $(2.7)$} & ${38.3}${\tiny $(4.9)$} & ${41.6}${\tiny $(3.6)$} & ${36.8}${\tiny $(5.6)$} &${43.4}${\tiny $(3.1)$}  & ${50.0}${\tiny $(6.0)$} & ${39.6}${\tiny $(4.4)$} & ${44.0}${\tiny $(4.0)$} & ${48.5}${\tiny $(2.7)$}   \\
        FSLS & & ${42.8}${\tiny $(8.0)$} & ${49.0}${\tiny $(10.5)$} &  ${53.4}${\tiny $(11.8)$} & ${52.7}${\tiny $(2.5)$}  &  ${62.2}${\tiny $(1.5)$} &  ${65.2}${\tiny $(2.7)$} & ${39.0}${\tiny $(5.5)$} & ${48.8}${\tiny $(1.7)$} & ${50.8}${\tiny $(3.1)$}   \\
        Unified Baseline & &${39.0}${\tiny $(6.1)$} & ${45.9}${\tiny $(9.4)$} & ${47.0}${\tiny $(8.3)$}  & ${52.8}${\tiny $(2.1)$} & ${60.8}${\tiny $(3.4)$} & ${60.0}${\tiny $(4.9)$} & ${37.6}${\tiny $(6.8)$} & ${45.9}${\tiny $(4.5)$} & ${47.8}${\tiny $(4.2)$}  \\
        \midrule
        $\_$ & \parbox[t]{1mm}{\multirow{6}{*}{\rotatebox[origin=c]{90}{L-TapNet}}} & ${42.6}$ {\tiny $(3.8)$} & ${50.8}${\tiny $(4.1)$} & ${50.8}${\tiny $(2.8)$} & ${53.2}${\tiny $(2.3)$} & ${63.3}${\tiny $(1.6)$} & ${68.5}${\tiny $(0.7)$} & ${44.5}${\tiny $(4.5)$} & ${52.3}${\tiny $(2.1)$} & ${52.5}${\tiny $(2.5)$}  \\
        Fine-tuning & & ${43.9}${\tiny $(11.4)$}& ${54.8}${\tiny $(9.4)$} & ${57.2}${\tiny $(5.0)$} & ${52.2}${\tiny $(3.2)$} &${64.4}${\tiny $(2.1)$}  & ${68.5}${\tiny $(0.7)$} & ${38.8}${\tiny $(3.7)$} & ${48.1}${\tiny $(2.5)$} & ${51.7}${\tiny $(3.6)$} \\
        CONTAINER & & ${34.4}${\tiny $(4.7)$} & ${43.6}${\tiny $(4.6)$} & ${45.3}${\tiny $(4.2)$}&${44.9}${\tiny $(10.8)$} & ${63.4}${\tiny $(2.8)$}& ${\textbf{69.4}}${\tiny $(1.1)$} & ${39.5}${\tiny $(4.6)$} & ${49.2}${\tiny $(4.7)$} & ${52.8}${\tiny $(3.3)$} \\
        L-TapNet &  & ${37.2}${\tiny $(4.6)$} & ${45.4}${\tiny $(2.8)$} & ${45.1}${\tiny $(3.7)$} & ${52.1}${\tiny $(2.2)$} & ${62.6}${\tiny $(2.6)$} & ${68.0}${\tiny $(1.4)$} & ${44.9}${\tiny $(5.4)$} & ${49.7}${\tiny $(2.9)$} & ${52.0}${\tiny $(5.2)$} \\
        FSLS & & $\underline{51.8}${\tiny $(6.4)$} & ${59.1}${\tiny $(6.3)$} & ${60.4}${\tiny $(6.7)$} & ${51.1}${\tiny $(10.2)$} & ${63.8}${\tiny $(2.2)$}  &  ${68.5}${\tiny $(1.6)$} & ${45.0}${\tiny $(5.6)$} & ${53.6}${\tiny $(3.1)$} & ${54.2}${\tiny $(2.2)$} \\
        Unified Baseline & &${45.8}${\tiny $(5.6)$} & ${52.7}${\tiny $(6.9)$} & ${59.4}${\tiny $(5.3)$}& $\textbf{56.1}${\tiny $(2.1)$} & ${63.6}${\tiny $(2.5)$} & ${68.0}${\tiny $(1.8)$}  & ${45.8}${\tiny $(4.6)$} & ${51.2}${\tiny $(2.9)$} & ${55.3}${\tiny $(2.2)$}  \\
        \midrule
        $\_$ & \parbox[t]{1mm}{\multirow{6}{*}{\rotatebox[origin=c]{90}{FSLS}}} & ${42.9}${\tiny $(4.0)$} & ${49.9}${\tiny $(4.3)$} & ${52.5}${\tiny $(2.7)$} & ${43.5}${\tiny $(4.9)$} & ${58.2}${\tiny $(1.1)$} &  ${64.1}${\tiny $(0.7)$} & ${46.1}${\tiny $(7.0)$} & ${49.3}${\tiny $(3.9)$} & ${53.5}${\tiny $(3.5)$}  \\
        Fine-tuning & & ${49.6}${\tiny $(5.2)$}  & ${56.0}${\tiny $(7.7)$}  & ${56.5}${\tiny $(6.5)$} & ${44.9}${\tiny $(5.0)$} & ${59.2}${\tiny $(2.0)$} &  ${64.2}${\tiny $(1.5)$} & ${39.1}${\tiny $(5.0)$} & ${45.7}${\tiny $(3.2)$} & ${51.3}${\tiny $(3.6)$} \\
        CONTAINER & & ${32.0}${\tiny $(4.5)$} & ${40.9}${\tiny $(4.1)$} & ${45.1}${\tiny $(3.8)$} & ${48.0}${\tiny $(1.6)$} & ${59.2}${\tiny $(3.2)$} & ${64.1}${\tiny $(2.5)$} & ${40.0}${\tiny $(3.6)$} & ${45.6}${\tiny $(4.6)$} & ${48.9}${\tiny $(4.5)$}  \\
        L-TapNet &  & ${36.8}${\tiny $(3.0)$} & ${43.3}${\tiny $(3.4)$}  & ${47.1}${\tiny $(2.7)$} &  ${43.9}${\tiny $(2.1)$}& ${55.9}${\tiny $(1.9)$}  & ${62.4}${\tiny $(1.5)$} & ${44.1}${\tiny $(4.6)$} & ${47.3}${\tiny $(3.1)$} & ${51.0}${\tiny $(2.7)$} \\
        FSLS & & ${51.7}${\tiny $(7.3)$} & ${\textbf{61.5}}${\tiny $(7.9)$} & ${\underline{66.2}}${\tiny $(4.3)$} & ${50.8}${\tiny $(1.9)$} & ${59.3}${\tiny $(1.9)$} & ${65.5}${\tiny $(1.4)$} & ${46.4}${\tiny $(3.4)$} & ${54.4}${\tiny $(3.5)$} & ${56.2}${\tiny $(2.2)$}   \\
        Unified Baseline & &${44.5}${\tiny $(8.5)$} & ${53.4}${\tiny $(7.2)$} & ${57.7}${\tiny $(6.4)$} & ${50.6}${\tiny $(3.3)$} & ${59.7}${\tiny $(0.7)$} & ${64.0}${\tiny $(0.8)$} & ${46.1}${\tiny $(4.4)$} & ${50.4}${\tiny $(4.4)$} & ${55.1}${\tiny $(2.1)$}  \\
        \midrule
        $\_$ & \parbox[t]{1mm}{\multirow{6}{*}{\rotatebox[origin=c]{90}{Unified Baseline}}} & ${47.4}${\tiny $(5.8)$} & ${55.9}${\tiny $(3.4)$}  & ${56.8}${\tiny $(3.4)$}  & ${49.1}${\tiny $(1.2)$} & ${63.9}${\tiny $(1.1)$} & ${68.2}${\tiny $(1.3)$} & $\textbf{51.7}${\tiny $(5.9)$} & $\textbf{57.1}${\tiny $(2.0)$} & ${56.8}${\tiny $(4.0)$}  \\
        Fine-tuning & & ${51.2}${\tiny $(4.8)$} &  ${58.6}${\tiny $(8.3)$} & ${61.9}${\tiny $(8.7)$} &${52.0}${\tiny $(1.1)$}   & ${63.6}${\tiny $(2.2)$}  & ${68.1}${\tiny $(1.4)$} & ${40.0}${\tiny $(5.9)$} & ${51.8}${\tiny $(4.5)$} & ${57.1}${\tiny $(3.4)$}  \\
        CONTAINER & & ${34.3}${\tiny $(3.5)$} & ${43.9}${\tiny $(4.9)$} & ${50.9}${\tiny $(3.1)$} & ${51.7}${\tiny $(2.0)$} & ${63.7}${\tiny $(1.4)$} & ${67.8}${\tiny $(1.5)$} & ${47.5}${\tiny $(4.6)$} & ${51.7}${\tiny $(3.7)$} & ${55.0}${\tiny $(2.9)$}  \\
        L-TapNet &  & ${42.3}${\tiny $(4.0)$} & ${49.0}${\tiny $(4.6)$} & ${51.6}${\tiny $(3.7)$} & ${49.1}${\tiny $(3.2)$} & ${63.5}${\tiny $(2.1)$} & ${67.5}${\tiny $(1.3)$} & ${47.2}${\tiny $(6.1)$} & ${53.4}${\tiny $(2.0)$} & ${55.0}${\tiny $(3.6)$}  \\
        FSLS & & ${\textbf{56.4}}${\tiny $(5.6)$} & ${\underline{61.4}}${\tiny $(6.7)$} & ${\textbf{67.3}}${\tiny $(4.2)$} & ${\underline{55.7}}${\tiny $(2.7)$} & ${\textbf{64.8}}${\tiny $(1.7)$} & ${68.9}${\tiny $(1.4)$} & $\underline{47.6}${\tiny $(4.1)$} & $\underline{57.1}${\tiny $(2.8)$} & $\textbf{58.6}${\tiny $(4.0)$}  \\
        Unified Baseline & & ${49.6}${\tiny $(6.5)$} & ${60.0}${\tiny $(6.0)$} & ${64.1}${\tiny $(7.2)$} & ${52.9}${\tiny $(3.3)$} & ${63.8}${\tiny $(2.6)$} & ${\underline{69.2}}${\tiny $(0.7)$} & ${45.4}${\tiny $(4.4)$} & ${53.5}${\tiny $(2.3)$} & $\underline{57.4}${\tiny $(3.8)$}   \\
        \bottomrule
        \end{tabular}
    \end{threeparttable}
    \label{tab:class-transfer-prototype}
\end{table*}

\end{document}